\documentclass{article}
\usepackage{log_2023}						

\usepackage{booktabs}						
\usepackage{multirow}						
\usepackage{amsfonts}						
\usepackage{graphicx}						
\usepackage{duckuments}						

\usepackage{graphicx}
\usepackage{subcaption}
\usepackage{algorithm}
\usepackage{algorithmic}
\usepackage{pifont}
\usepackage[numbers,compress,sort]{natbib}	

\title[]{Multicoated and Folded Graph Neural Networks with \\ Strong Lottery Tickets}

\author[J. Yan et al.]{%
  Jiale Yan, \space  Hiroaki Ito, \space  Ángel López García-Arias, \space  Yasuyuki Okoshi, \space  Hikari Otsuka, \\ \space  \textbf{Kazushi Kawamura,}  \space \textbf{Thiem Van Chu,}  \space \textbf{Masato Motomura\thanks{Corresponding author. Project URL: https://github.com/LouiValley/SLT-GNN}}
 \\
Tokyo Institute of Technology, Tokyo, Japan \\
\email\{yan.jiale, ito.hiroaki, lopez, okoshi.yasuyuki, otsuka.hikari, kawamura, thiem, motomura\}@artic.iir.titech.ac.jp}
\newcommand{\bhline}[1]{\noalign{\hrule height #1}}

\begin{document}

\maketitle

\begin{abstract}
The Strong Lottery Ticket Hypothesis (SLTH) demonstrates the existence of high-performing subnetworks within a randomly initialized model, discoverable through pruning a convolutional neural network (CNN) without any weight training. A recent study, called Untrained GNNs Tickets (UGT), expanded SLTH from CNNs to shallow graph neural networks (GNNs). However, discrepancies persist when comparing baseline models with learned dense weights. Additionally, there remains an unexplored area in applying SLTH to deeper GNNs, which, despite delivering improved accuracy with additional layers, suffer from excessive memory requirements. To address these challenges, this work utilizes Multicoated Supermasks (M-Sup), a scalar pruning mask method, and implements it in GNNs by proposing a strategy for setting its pruning thresholds adaptively. In the context of deep GNNs, this research uncovers the existence of untrained recurrent networks, which exhibit performance on par with their trained feed-forward counterparts. This paper also introduces the Multi-Stage Folding and Unshared Masks methods to expand the search space in terms of both architecture and parameters. Through the evaluation of various datasets, including the Open Graph Benchmark (OGB), this work establishes a triple-win scenario for SLTH-based GNNs: by achieving high sparsity, competitive performance, and high memory efficiency with up to 98.7\% reduction, it demonstrates suitability for energy-efficient graph processing.

\end{abstract}

\section{Introduction}
 Graph Neural Networks (GNNs)~\cite{GCN, GAT, GIN, li2019deepgcns,li2020deepergcn} have emerged as powerful models for graph-based learning tasks. Within this field, both shallow GNNs~\cite{GCN, GAT, GIN} and deep GNNs~\cite{li2019deepgcns,li2020deepergcn} hold significant presence, each characterized by their distinct layers and intricate structural complexities. Their success is attributed to the synergistic combination of the neighborhood aggregation scheme and weight sharing. The neighborhood aggregation scheme effectively captures local graph structures and inter-dependencies, while weight sharing enables GNNs to generalize across nodes, extracting meaningful features from large-scale graphs. Unfortunately, computational overheads increase rapidly when fitting GNNs to large-scale graphs. 

To decrease the computational burden, studies~\cite{chen2021unified,wang2022searching,hui2023rethinking, UGT_LOG'22} have utilized the Lottery Ticket Hypothesis (LTH)~\cite{lottery_theory,edgepopup} to identify graph lottery tickets, sparse subnetworks extracted from dense GNNs that can perform comparably to the original models. Among them, ~\cite{chen2021unified,wang2022searching,hui2023rethinking} simultaneously simplify the input graph and prune GNNs' weights. These approaches belong to the Weak Lottery Ticket Hypothesis (WLTH)~\cite{you2019drawing,lee2019signal, wang2020pruning, wang2019picking, frankle2020pruning, sreenivasan2022rare}, as they require weight re-training. On the other hand, the Untrained GNNs Tickets (UGT) study~\cite{UGT_LOG'22} focuses on pruning GNNs' weights and does not require any subsequent weight training, adhering to the Strong Lottery Ticket Hypothesis (SLTH)~\cite{zhou2019deconstructing,malach2020proving,pensia2020optimal,orseau2020logarithmic,okoshi2022multicoated}. This paper primarily focuses on SLTH due to its potential as a hardware-friendly algorithm for GNNs. For one thing, SLTH allows transforming entire GNN computations into Sparse Matrix-Matrix Multiplication (SpMM), potentially paving the way for a paradigm shift that eradicates the need for GNN's hybrid architectures~\cite{yan2020hygcn,yoo2023sgcn}, which utilizes General Matrix-Matrix Multiplication (GeMM) and SpMM engines. For another, SLTH can be leveraged for an efficient SLTH inference hardware implementation~\cite{hirose2022hiddenite}, where random weights are generated on the fly instead of stored in memories. Nevertheless, the application of SLTH in GNNs introduces two challenges:

\begin{enumerate}

\item For shallow GNNs such as GCN, GAT, and GIN, UGT~\cite{UGT_LOG'22} provides better accuracy than Edge-Popup~\cite{edgepopup}. When observing the sparsity value of the weights varies from 0\% to nearly 100\%, we find discrepancies persist when compared to baseline models trained using dense weights. There is still a question for research: \textbf{Is it possible to devise a better SLTH that maintains high accuracy for shallow GNNs?}

\item Concerning deep GNNs, UGT explores the implementation of SLTH on GCN, GAT, and GIN by directly increasing the number of layers. However, a limitation is that these enlarged models do not match the accuracy level of two-layer models. Concurrently, ResGCNs~\cite{li2019deepgcns,li2020deepergcn} exhibit improved accuracy with increased layer depth but need more memory consumption. Although ~\cite{wang2022searching} explored  WLTH on ResGCN+~\cite{li2020deepergcn}, the SLTH has never been explored for them. Here are two questions: \textbf{First, is applying SLTH to deep GNNs feasible? Second, can we further enhance model efficiency by reducing the size without compromising the high accuracy?}

\end{enumerate}

Inspired by recent works emerging around the SLTH, including multicoated supermasks (M-Sup)~\cite{okoshi2022multicoated} and folding methods~\cite{lopez'access,Garcia-AriasHMY21}, this work exploits the SLTH on GNNs by implementing M-Sup on GNNs and analyzes their effects. Furthermore, it adopts the folding method to optimize deep GNNs such as ResGCNs. The contributions of this paper are summarized as follows:

\begin{enumerate}

\item This research identifies a significant performance improvement in GNNs using M-Sup, i.e., with solely using randomly initialized weights. By identifying high-performing subnetworks in their random initial states, this work can surpass the performance of single-coated supermask (S-Sup), e.g., UGT, and the accuracy of the dense-weight training baseline. Contrary to the WLTH, this process does not need weight learning.

\item Based on M-Sup, the research analyzes the weight score distribution of GNN models and proposes an adaptive strategy to determine pruning thresholds for different score distributions. Experiments show its advantage in terms of accuracy.

\item This paper is the first time to demonstrate the existence of untrained recurrent graph subnetworks within deep GNNs, exhibiting comparable performance to their trained feed-forward counterparts. Furthermore, this study introduces the Multi-stage folding (MSF) and Unshared Mask methods designed for a broader search space within the domains of network structure and weight scores. By capitalizing on the signed Kaiming Constant (SC) initialization without weight training, ResGCN+ and DyResGEN models~\cite{li2020deepergcn} achieve memory reductions of 72\% and 98.7\%, respectively, while maintaining an accuracy comparable to the baseline models. 
\end{enumerate}
\section{Related Work}
\subsection{Graph Neural Networks} 
GNNs have demonstrated state-of-the-art performance on graph-structure tasks~\cite{GCN, GAT, GIN, hamilton2017inductive, corso2020principal, ying2018jiaxuan}. After GCN~\cite{GCN} proposed a two-layer GNN for node classification, more GNNs such as GAT~\cite{GAT}, GIN~\cite{GIN}, GraphSage~\cite{hamilton2017inductive}, PNA~\cite{corso2020principal}, and DiffPool~\cite{ying2018jiaxuan} were put forward. Given an undirected graph $\mathcal{G}  = \left \{ \mathcal{V}, \mathcal{E}  \right \} $, where ${\mathcal{V}}$ is a set of nodes and $\mathcal{E}$ is a set of edges, there are $|\mathcal{V}|$ nodes, and each node has a feature vector $x_i\in \mathbb{R}^F$, where $F$ is the number of a node feature. 
The adjacency matrix is $A\in \mathbb{R} ^{\left | \mathcal{V}\times \mathcal{V} \right | }$ and node features are $X\in \mathbb{R} ^{\left | \mathcal{V} \right |\times F}$. A typical 2-layer GNN can be defined as
\begin{align}
\mathcal{G}^{(2)}{\left (  \hat{\mathit{A}}  ,  \mathit{X}   ,\mathit{W}    \right ) } = \mathrm{softmax}  \left ( \hat{A}  \sigma \left ( \hat{A}XW^{(0)} \right )  W^{(1)} \right ),
\end{align}
where $W^{(0)}$ and $W^{(1)}$ are learnable weights for two layers, $\hat{A} =\tilde{D}^{-\tfrac{1}{2} } \left ( A+I \right )\tilde{D}^{-\tfrac{1}{2} }$ is the normalized adjacency matrix, and $\tilde{D}$ is the degree matrix of $\hat{A}$. In addition, $\mathrm{softmax}$ is the softmax function, and $\sigma\left ( \cdot  \right )$ is an activation function.

Despite GNNs having seen rapid and substantial progress, most prior studies still employ shallow structures. Unlike CNNs, GNNs suffer from vanishing gradient and over-smoothing~\cite{li2018deeper} when going deeper. Inspired by the benefit of training deep CNN-based networks~\cite{he2016deep, yu2015multi}, ResGCN variants~\cite{li2019deepgcns, li2020deepergcn} were developed by adopting residual connections and dilated convolutions to GCNs. They solve the degradation problem and outperform traditional shallow GNNs. Moreover, the development of deep GNNs has also benefited from techniques like dropping DropEdge~\cite{dropedge} and DropNode~\cite{dropnode}, as well as normalization methods~\cite{nodenorm, zhao2019pairnorm}.


\subsection{Weak and Strong Lottery Ticket Hypothesis}

A breakthrough paper~\cite{lottery_theory} revealed Lottery Ticket Hypothesis (LTH), which showed that an overparameterized neural network contains a subnetwork that can be trained in isolation to match the original models. Similar studies have quickly developed into two groups: Weak Lottery Ticket Hypothesis (WLTH) and Strong Lottery Ticket Hypothesis (SLTH). For WLTH, the subnetworks need iterative training, pruning, and resetting the remaining weights to their original value. After LTH has shown its potential capability, other studies focused on the early training stage~\cite{you2019drawing} or even before training~\cite{lee2019signal, wang2020pruning, wang2019picking, frankle2020pruning, sreenivasan2022rare}. Not limited to CNNs, GLT~\cite{chen2021unified} was the first study to generalize WLTH to GNNs. Later, DGLT~\cite{wang2022searching} explored transferring a random ticket to a graph lottery ticket. GLT~\cite{hui2023rethinking} proposed a new auxiliary loss function to guide pruning better when the graph sparsity is high. On the other hand, the SLTH~\cite{zhou2019deconstructing} revealed that an overparameterized network has high-performing subnetworks at a randomly initialized state, only needing pruning to be discovered. After that, a series of papers~\cite{malach2020proving,pensia2020optimal,orseau2020logarithmic,okoshi2022multicoated} added theoretical proof to the theory on CNNs. For the GNN field, distinct from~\cite{chen2021unified,wang2022searching,hui2023rethinking}, UGT~\cite{UGT_LOG'22} achieved competitive accuracy without any model weight training.

\subsection{Hidden-Fold Network}
The folding method can convert deep feed-forward networks into shallow recurrent models, which is applied in ResNets~\cite{he2016deep}. A range of studies~\cite{greff2016highway, liao2016bridging, lopez'access} showed that ResNets can be closely approximated as unrolled shallow recursive neural networks, attributing the benefits of additional layers to recursive iterations. In particular, ~\cite{liao2016bridging} demonstrated that the accuracy is only marginally impacted when folding ResNet into a 4-layer model. Building on this concept of folded networks, ~\cite{Garcia-AriasHMY21} pioneered the exploration of SLTH in folded residual networks. This approach was subsequently enhanced by \cite{lopez'access} without causing a significant loss. A blend of SLTH with folded networks is called Hidden-Fold Network (HFN)\cite{Garcia-AriasHMY21, lopez'access}.



\section{Multicoated and Folded GNNs}

\subsection{Single-coated Supermask (S-Sup) in strong GNN tickets}
UGT~\cite{UGT_LOG'22} utilizes Edge-Popup~\cite{edgepopup} method by applying S-Sup to acquire strong tickets from a GNN. Given a $\mathcal{G\left ( \hat{\mathit{A}}, \mathit{X}, \mathit{W}\right ) }$, UGT formulates the weight matrix of layer $l$ as follows
\begin{align}
W^{(l)}= W^{(l)}_{\mathrm{rand}} \odot \mathcal{H}(\mathcal{S}^{(l)}), \mathcal{H}(\mathcal{S}^{(l)})\in \left \{  0, 1\right \},
\end{align}

where $\odot$ signifies the Hadamard product operator, $W^{(l)}_{\mathrm{rand}}$ denotes random tensors for weights, $\mathcal{H}$ is the binary supermask generation function, and $\mathcal{S}^{(l)}$ is the underlying score for a specific layer. Each element $h(s_{uv})$ in $\mathcal{H}(\mathcal{S}^{(l)})$ is defined as 
\begin{align}
h(s_{uv})=\begin{cases}
 1 &,  \left | s_{uv} \right | \ge s_{\mathrm{threshold} }  \\
 0 &,  \left | s_{uv} \right | <  s_{\mathrm{threshold} } 
\end{cases} ,
\end{align}
where the $s_{uv}$ is the score corresponding to a weight value in $W^{(l)} \in \mathbb{R} ^{\left |  {U}\times {V} \right |}$ and $s_{\mathrm{threshold} }$ is a threshold. The $s_{\mathrm{threshold}}$ is based on a predetermined sparsity value $k\%$. This value aids in pruning the smallest $k\%$ of the scores $\left | s_{uv} \right | \in \mathcal{S}^{(l)}$.






\subsection{M-Sup with adaptive threshold for GNNs}

\begin{figure}[tb]
    \centering
    \begin{subfigure}{0.63\textwidth}
        \includegraphics[width=\textwidth]{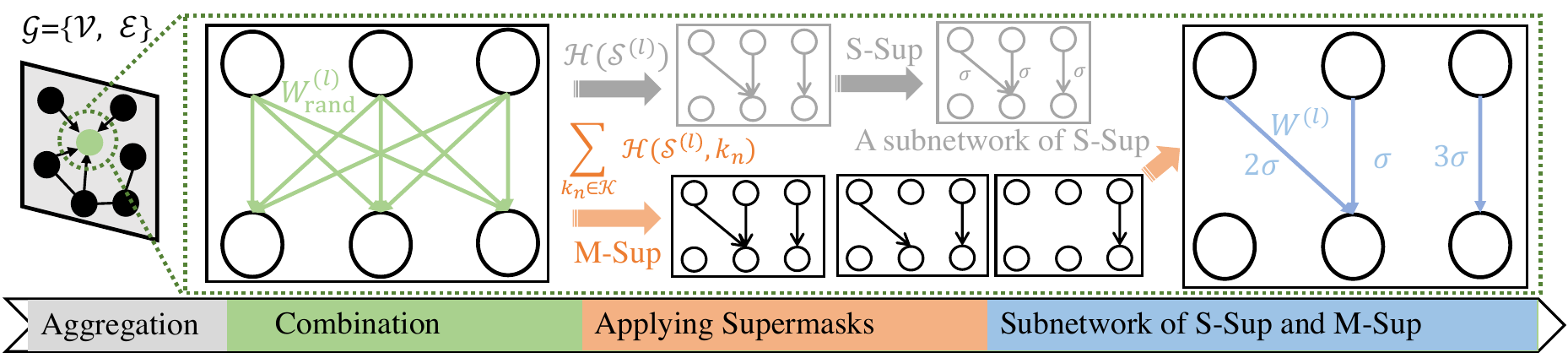}
        \caption{}
        \label{fig:GNNweightdis_image1}
    \end{subfigure}
    \begin{subfigure}{0.35\textwidth}
        \includegraphics[width=\textwidth]{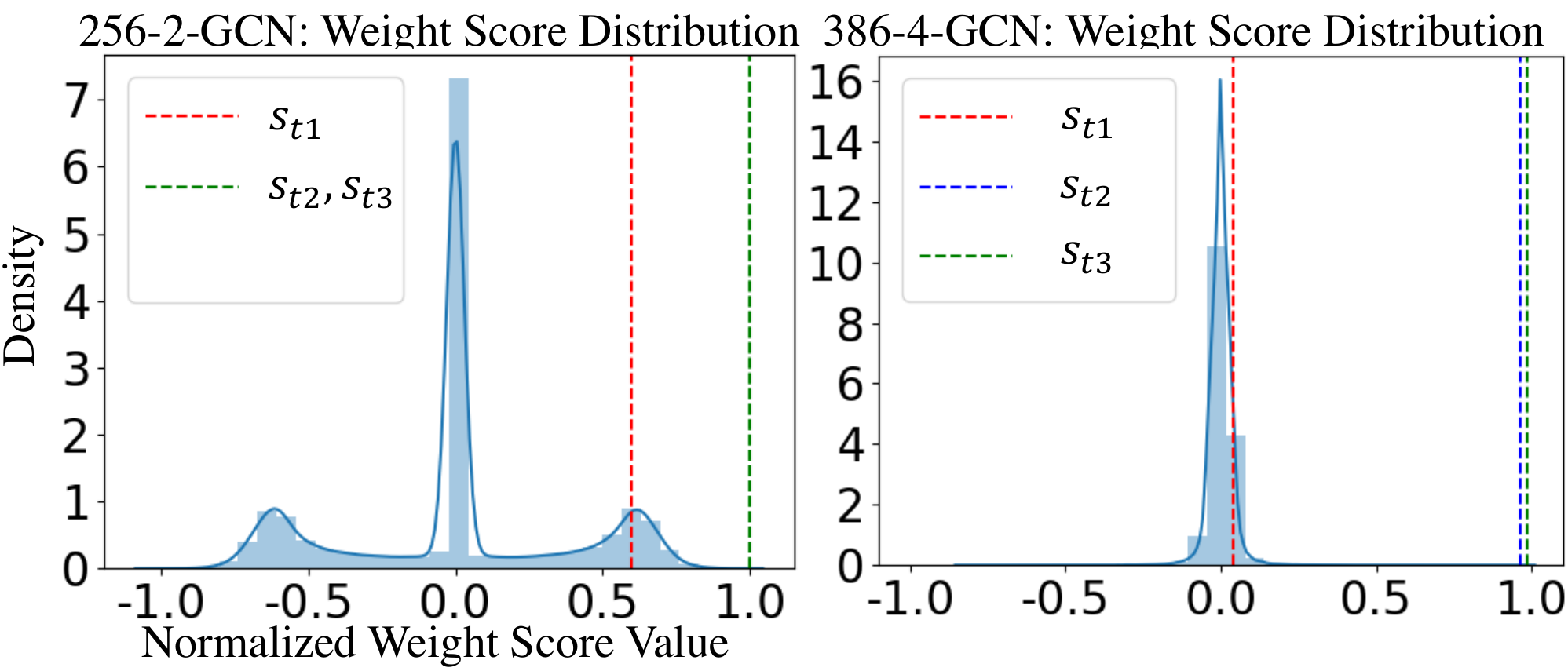}
        \caption{}
        \label{fig:GNNweightdis_image2}
    \end{subfigure}
    \caption{A GNN with the M-Sup method: (a) shows how to apply M-Sup to a GNN. Weights are randomly initialized to $\pm \delta $, where $\delta$ is the standard deviation of the Kaiming normal distribution. (b) illustrates weight score distributions of GNN models. 256-2-GCN has 256 hidden neurons and two layers used on Cora, and 386-4-GCN has 386 hidden neurons and four layers used on OGBN-Arxiv. $s_{t_1},s_{t_2},s_{t_3}$ are the thresholds with the Linear method.}
    \vspace*{-0.4cm}
    \label{fig:GNNweightdis}
\end{figure}

Building on the prior work~\cite{okoshi2022multicoated}, which applies M-Sup to ResNet, this study expands M-Sup to GNNs without requiring weight training. After applying one supermask with multiple coats, the weight matrix of layer $l$ becomes
\begin{align}
W^{(l)} = W^{(l)}_{\mathrm{rand}} \odot \sum_{k_n\in \mathcal{K} }^{} \mathcal{H}(\mathcal{S}^{(l)},k_n), \mathcal{H}(\mathcal{S}^{(l)},k_n)\in \left \{  0, 1\right \}.
\end{align}
$\mathcal{K}$ is a set of sparsity values, corresponding to $N$ coats, where $k_1\le k_2 \le ... \le k_N$. Here, the latter coat is more sparse than the previous one. Each value $k_n$ in the set $\mathcal{K}$ is determined by threshold $s_{t_n}$, which results in pruning $k_n\%$ of $\mathcal{S}^{(l)}$ based on $\left | s_{uv} \right | < s_{t_n}$. This process is illustrated in \figurename~\ref{fig:GNNweightdis}.

 \textbf{Adaptive threshold.} This study proposes an adaptive method to define $\mathcal{K}$ for M-Sup. ~\cite{okoshi2022multicoated} proposed Linear and Uniform methods to set the sparsity $\mathcal{K}$. The Linear method defines $s_{t_n} = s_{t_1} + 3\sigma_s \times \frac{n-1}{N}$, where $s_{t_1}$ is the threshold score corresponding to sparsity $k_1$, and $\sigma_s$ refers to the standard deviation of the normalized weight scores. The Uniform method directly defines ${k_n} = k_1 + (100\%-k_1)\times \frac{n-1}{N}$. In a practical implementation, the Linear method proved less effective. As depicted in \figurename~\ref{fig:GNNweightdis_image2}, the weight score distribution for a 385-4-GCN model is suitable for the Linear method, with three effective values $s_{t_n} < 1.0$, but for a 256-2-GCN model, the weight score distribution becomes multimodal, presenting a challenge. In this specific case, both $s_{t_2}$ and $s_{t_3}$ equate to 1.0, making the respective coats ineffective. The situation will further deteriorate if UGT's linear decay schedule~\cite{UGT_LOG'22} is applied, where sparsity list are $\mathcal{K}_{t} = \mathcal{K}\cdot \frac{\beta \cdot t_{\mathrm{current\,epoch}}}{T_{\mathrm{total\,epochs}}}$, and $\beta$ is an empirical hyperparamete. These meaningless coats disrupt training by pruning partial weights since they do not begin at 100\% in the training phase, leading to potential problems in the model's performance.

Therefore, for the Linear threshold, an adaptive determination for $s_{t_n}$ corresponding to ${k_{n}}\in \mathcal{K}$ is proposed as follows:
\begin{align}
s_{t_n} = \left \{  s_{t_1} + \frac{3\sigma_s}{N} \times (n-1), \phi\right \}  \text{, Linear (w/ pre-training)}  
\end{align}
The method works as follows: when provided with a pre-trained single-coat model, it utilizes the Linear method to calculate $s_{t_n}$, but only if $s_{t_n}$ is smaller than $\alpha$. $\alpha$ is an empirical hyperparameter with a value of 0.9996. Otherwise, it is defined as $\phi$, which means the coat is invalid. If no pre-trained models are available, the method uses a uniform approach to directly calculate ${k_n}$.




\begin{figure}[t]
\centering
	\includegraphics[width=0.9\linewidth]{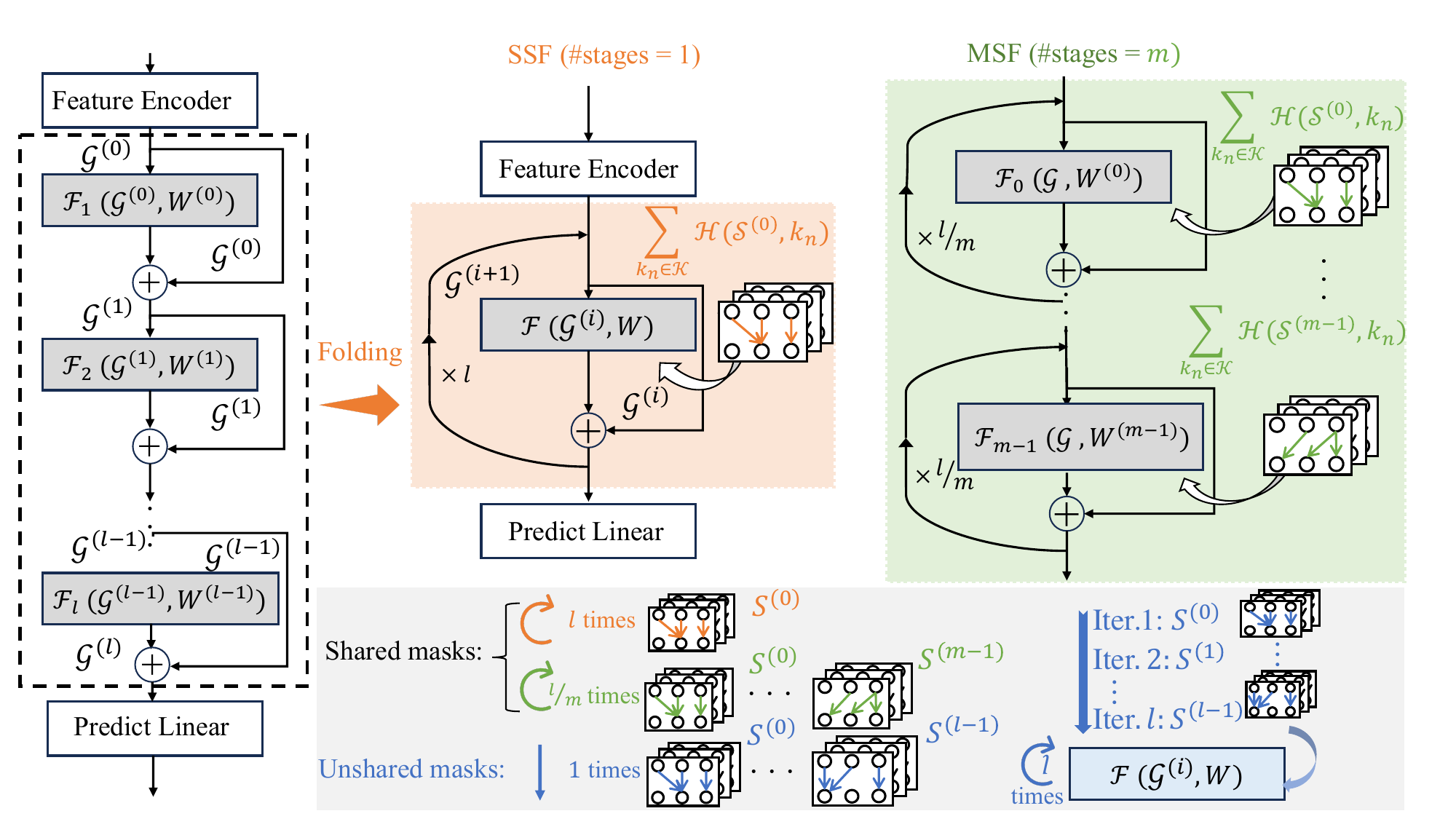}
	\caption{Folded GNN: a ResGNN can be folded into a recursive structure like a 1-stage or an n-stage MSF. Shared masks have a set for each stage and are reused as often as the number of iterations for one stage. Unshared masks are used only once, with individual masks for each iteration.}
    \vspace*{-0.35cm}  
	\label{fig:hidden-GNN}
\end{figure}

\subsection{Folded GNN}
\textbf{Hidden-Folded GNN.} Based on typical deep GNNs such as ResGCNs~\cite{li2019deepgcns,li2020deepergcn}, this work folds residual blocks into a single recurrent block. This transformation is represented in \figurename~\ref{fig:hidden-GNN}. This study reuses a general graph convolution operation $\mathcal{F}$~\cite{li2019deepgcns} with learnable parameters, including weights. Based on the folding structure, this paper also implements S-Sup and M-Sup. The corresponding formulas are as follows:
\begin{align}
\left\{\begin{matrix} \mathcal{G^\mathrm{(1)}} =  \mathcal{F}\left ( \mathcal{G^{\mathrm{(0)}}}, \mathcal{W}^\mathrm{(0)} \right ) + \mathcal{G^{\mathrm{(0)}}}
 \\
:
 \\ 
\mathcal{G}{^{(l)}} = \mathcal{F}\left ( \mathcal{G}^{(l-1)}, \mathcal{W}^{(l-1)} \right ) + \mathcal{G}^{(l-1)} 
\end{matrix} \right.
, \mathcal{W}{^{(l-1)}} = \cdots =\mathcal{W}{^{(0)} =  \mathcal{W}^{(0)}_{\mathrm{rand}} \odot \sum_{k_n\in \mathcal{K} }^{} \mathcal{H}(\mathcal{S}^{(0)},k_n)}.
\end{align}

Furthermore, the multi-folding stages (MFS) and Unshared Masks methods are proposed to explore optimized hidden-folded GNNs. The former expands multi-folding stages from the model architecture perspective and the latter from the weight scores perspective. They are outlined in Algorithm~\ref{alg:hidden-folded} and can be implemented individually or combined - e.g., MSF structures can be used in conjunction with unshared masks. The pseudocode for applying SLTH with the proposed methods is provided in Appendix~\ref{append:pseudocode}.

\textbf{MSF method.} In this approach, continuous stages are folded into multi-stage $m$, producing $m$ graph mappings denoted as $ \left \{  \mathcal{F_\mathrm{0} }, \mathcal{F_\mathrm{1}},\dots, \mathcal{F_\mathrm{m-1}}\right \} $. Given an $l$-layer and $m$-stages, each graph mapping block is utilized $\frac{l}{m}$ times. If $l$ is not a multiple of $m$, the integral division part of the blocks is reused, and the remainder part of the blocks is left unfolded. When $m=1$, it is single-stage folding (SSF). 

\textbf{Shared and Unshared Supermasks.} In contrast to shared supermasks, unshared supermasks are individually applied to each iteration. SSF structures with shared supermasks, a set of supermasks $\sum_{k_n\in \mathcal{K} }^{} \mathcal{H}(\mathcal{S}^{(0)},k_n)$, is shared across each iteration. In the unshared approach, $l$ sets of scores are created, denoted as $\sum_{k_n\in \mathcal{K} }^{} \mathcal{H}(\mathcal{S}^{(0)},k_n) \cdots \sum_{k_n\in \mathcal{K} }^{} \mathcal{H}(\mathcal{S}^{(l-1)},k_n)$, which are then deployed individually in each iteration.



\begin{algorithm}[tb]
	\renewcommand{\algorithmicrequire}{\textbf{Input:}}
	\renewcommand{\algorithmicensure}{\textbf{Output:}}
	\caption{Initialization and Inference about Folded GNN} 
	\label{alg:hidden-folded}
	\begin{algorithmic}[1]
		\STATE \textbf{Input}: $l$-layer , $m$-stages , iterations $r = \frac{l}{m}$, sparsity list $\mathcal{K}={k_1,...,k_N}$ \\ 
        \textcolor{gray}{\#If enable SSF, $m=1$; otherwise, $m>1$ for MSF.}
        \STATE \textbf{Output}: Features of the last layer: $\mathcal{G}^{l}$
            \IF{Enable Unshared Masks}
                \STATE Create $l$ sets of scores: $\mathcal{S} = \left \{\mathcal{S}^{(0)},\mathcal{S}^{(1)},\cdots, \mathcal{S}^{(l-1)} \right \}$  
            \ELSE
                \STATE Create $m$ sets of scores: $\mathcal{S} = \left \{\mathcal{S}^{(0)}, \mathcal{S}^{(1)},\cdot, \mathcal{S}^{(m-1)} \right \}$
            \ENDIF
        \FOR{$i=0$ to $l-1$} 
            \STATE Calculate the index of a reused layer: $j=i//r$
            \IF{Enable Unshared Masks}
            \STATE $\mathcal{W}{^{(i)} =  \mathcal{W}^{(j)}_{\mathrm{rand}} \odot \sum_{k_n\in \mathcal{K} }^{} \mathcal{H}(\mathcal{S}^{(i)},k_n)}$ 
            \ELSE
            \STATE $\mathcal{W}{^{(i)} =  \mathcal{W}^{(j)}_{\mathrm{rand}} \odot \sum_{k_n\in \mathcal{K} }^{} \mathcal{H}(\mathcal{S}^{(j)},k_n)}$ 
            \ENDIF
            \STATE Calculate the output of the current layer: $\mathcal{G}^{i+1}= \mathcal{F}_{j} \left ( \mathcal{G}^{(i)}, \mathcal{W}^{(i)} \right ) $
        \ENDFOR
        \STATE Return $\mathcal{G}^{l}$
	\end{algorithmic}  
\end{algorithm}

\section{Experiments and Discussions} 
This section describes experiments using various GNN architectures and datasets to assess the effectiveness of the proposed methods.

\textbf{GNN Architectures.} For shallow GNNs, this study utilizes GCN~\cite{GCN}, GAT~\cite{GAT}, and GIN~\cite{GIN}. For deep GNNs, ResGCNs~\cite{li2020deepergcn} are used, including a 7-layer DyResGEN and a 28-layer ResGCN+.

\textbf{Datasets.} The experiments uses three widely used graph datasets, Cora, Citeseer, and PubMed~\cite{GCN}, and the large-scale graph dataset OGBN-Arxiv~\cite{hu2020open} for node-level tasks (metric: Accuracy\%). Additionally, the research adopts OGBG-Molhiv and OGBG-Molbace~\cite{hu2020open} for graph-level tasks (metric: ROC-AUC\%). Node-level tasks involve predicting properties or characteristics at the individual node level within a graph. Graph-level tasks focus on predictions about the graph structure.

\textbf{Experimental settings.} During training, this study employs two weight initialization methods as~\cite{edgepopup}: Signed Kaiming Constant (SC) and Kaiming Normal (KN). Both methods utilize a scaling factor of $\sqrt{1/(1 - k_1)}$, defined according to the sparsity parameter $k_1$. Kaiming Uniform (KU) initialization is applied to weight scores. Through the experiments, SC performs better than KN in the S-Sup method, consistent with results reported in~\cite{okoshi2022multicoated}. Detailed findings are in Appendix~\ref{append:SCKN}. Since the SC is well-suited to a specialized SLTH inference accelerator~\cite{hirose2022hiddenite}, we prefer using the SC method. For graph-level tasks, the weights of an embedding list, denoted by $\left \{  W_1\in \mathbb{R} ^ {N_1\times M},\cdots, W_n\in \mathbb{R} ^ {N_n\times M} \right \}$, are also initialized with the SC method. We set the weight scores as a vector in $\mathbb{R} ^ {1\times M}$ rather than the same size of weights. It is applied to the look-up table's final output row by row, resulting in a memory-efficient compression. For sparsity scheduling, the experiments adopt a linear decay schedule as UGT~\cite{UGT_LOG'22}. Further implementation details are provided in Appendix~\ref{append:implementationde}. 

\textbf{Model Compression Scheme.} This article discusses the memory sizes of models, taking into account a particular compression method tailored for specialized hardware. Weights and biases are considered to occupy 32 bits each. However, it is not required to store the weights if they are initialized in an SC manner, as they can be generated on the fly from the original seed using a random number generator. Additionally, this seed can be replaced with a hash of other model parameters~\cite{hirose2022hiddenite}, eliminating the need to store it. As a result, the memory requirement for models employing supermask training consists solely of the supermask's size, equating to one bit per weight for the S-Sup and $(1+\sum_{n=1}^{N-1}k_n )$ bits per weight averagely for M-Sup with N coats, in addition to the affine normalization parameters, bias and other things that a model may use. 



\subsection{M-Sup finds better graph lottery tickets than S-Sup}

This subsection evaluates shallow and deep GNNs in node- and graph-level tasks. As the weight sparsity varies, the findings show the distinct advantage of M-Sup over S-Sup while achieving comparable accuracy as the baseline models, which use dense-weight learning (DWL).

\textbf{Shallow GNNs in node- and graph-level tasks.} For node-level tasks, we adopt 2-layer GNNs on Cora, Citeseer, and PubMed datasets, and 4-layer GNNs with affine batch normalization (BN) on OGBN-Arxiv. For graph-level tasks, we use a 3-layer GCN with embedding lists on OGBG-Molhiv and OGBG-Molbace. 

\begin{figure}[h]
	\centering
	\includegraphics[width=0.65\linewidth]{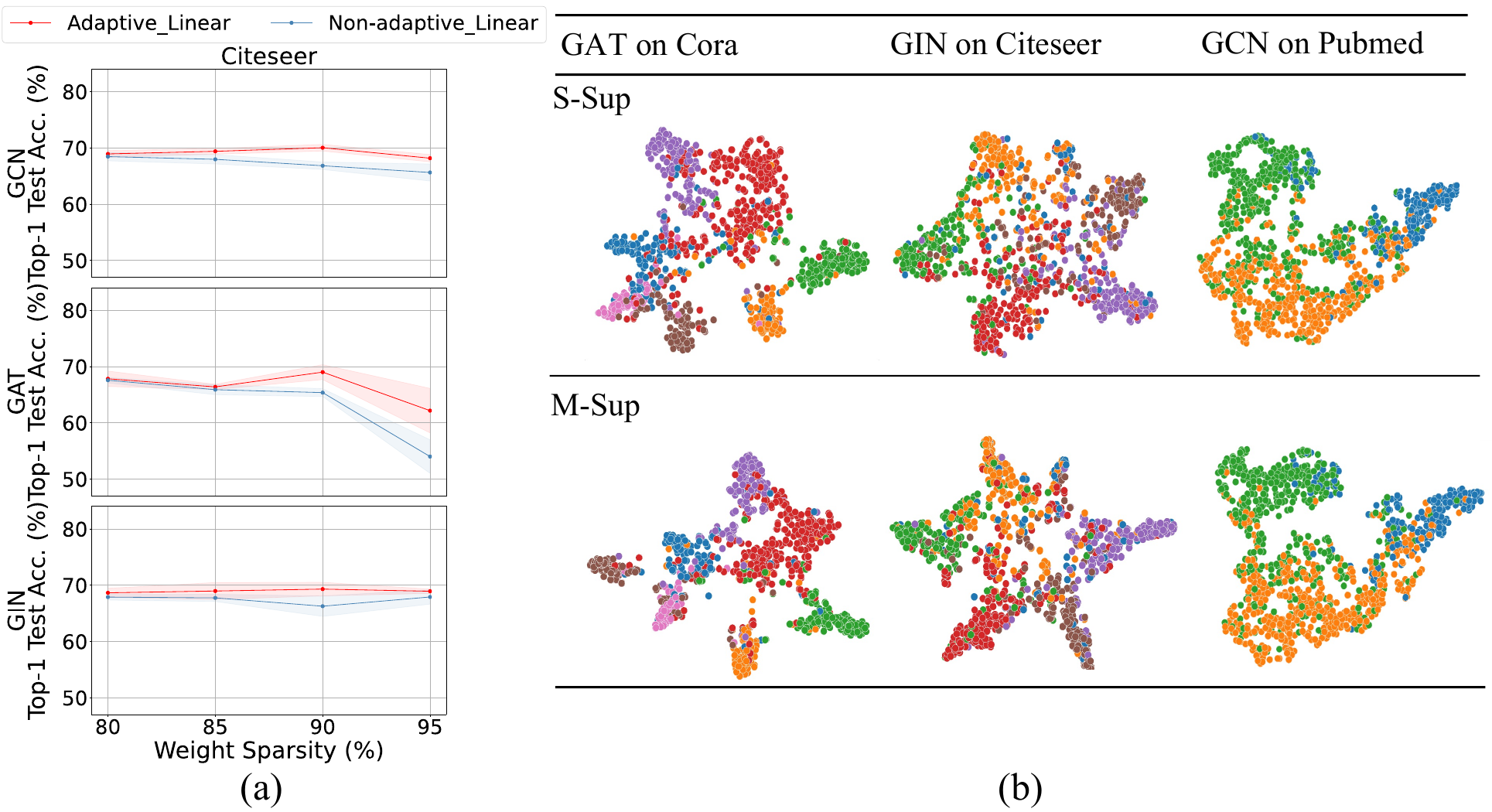}
	\caption{Advantages of adaptive Linear threshold and M-Sup methods: (a) shows the difference in accuracy between adaptive Linear threshold and no-adaptive Linear threshold for 3 GNN models on Citeseer. (b) shows the T-Distributed Stochastic Neighbor Embedding (TSNE) visualization of node representations learned by S-Sup and M-Sup (sparsity is 15\%).}
	\label{fig:tsne-gnn}
    \vspace*{-0.35cm}  
\end{figure}

Figure~\ref{fig:tsne-gnn} (a) shows the difference in accuracy between using an adaptive Linear threshold ($\alpha=0.9996$) and a non-adaptive Linear threshold ($\alpha=1.0$). Experiments show the adaptive Linear threshold is better when sparsity $ \ge 80\%$, because some ineffective coats appear in a non-adaptive Linear threshold. Other GNNs' experiment results are provided in Appendix~\ref{append:adaptive}. The adaptive Linear threshold is used in the following experiments.

\begin{figure}[h]
	\centering
	\includegraphics[width=1.0\linewidth]{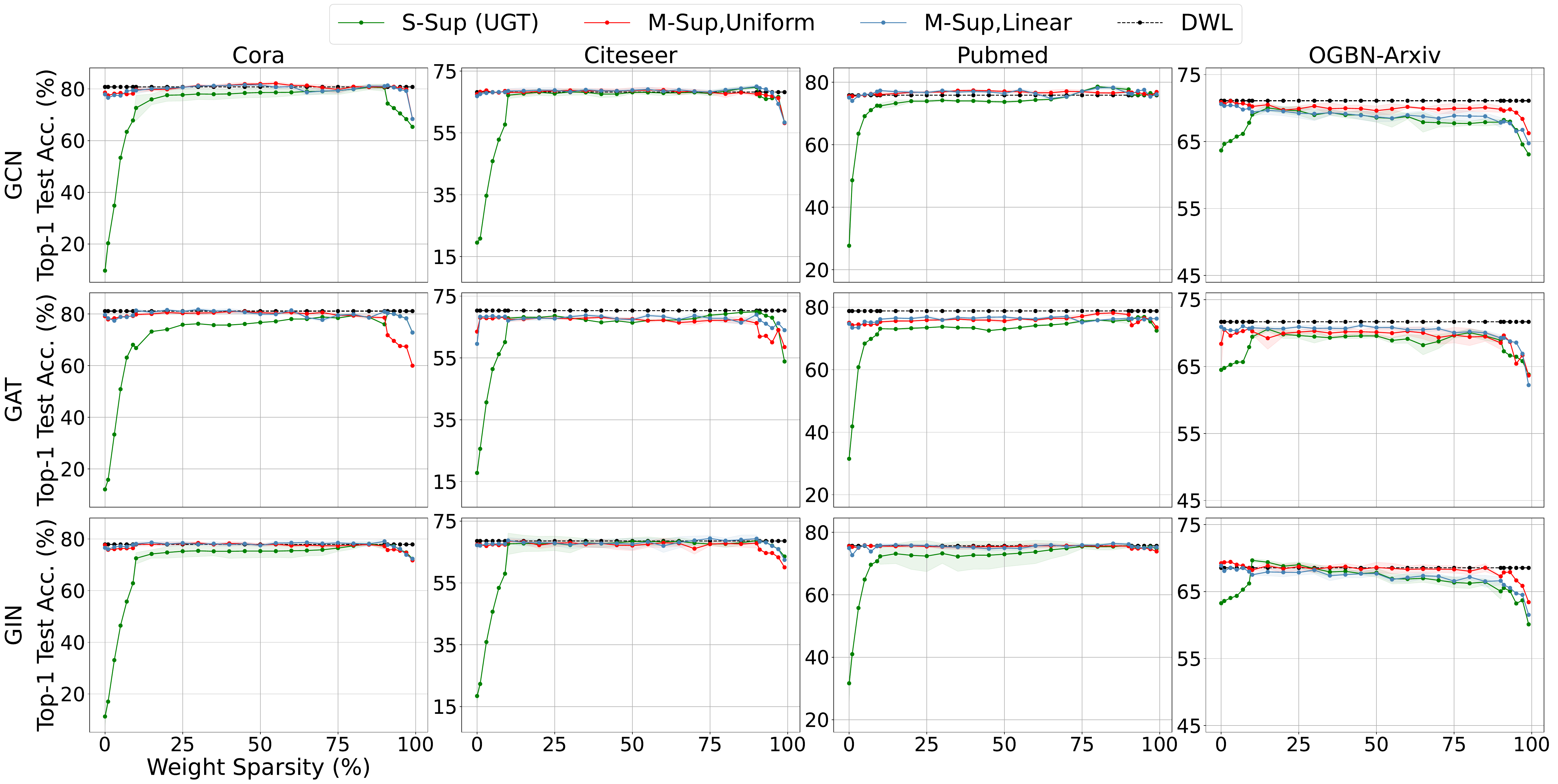}
	\caption{The accuracy of GCN, GAT, and GIN with various sparsity values on four datasets: experiments use 2-layer GNNs of width 256 for the Cora, Citeseer, and Pubmed datasets. For the OGBN-Arxiv datasets, a 4-layer GNN with a width of 386 is used, and affine BN is applied. The head of GAT is 1 for the following experiments.}
	\label{fig:shallow-GCN-nodel-level-tasks}
    \vspace*{-0.3cm}  
\end{figure}

\begin{figure}[h]
	\centering
	\includegraphics[width=1.0\linewidth]{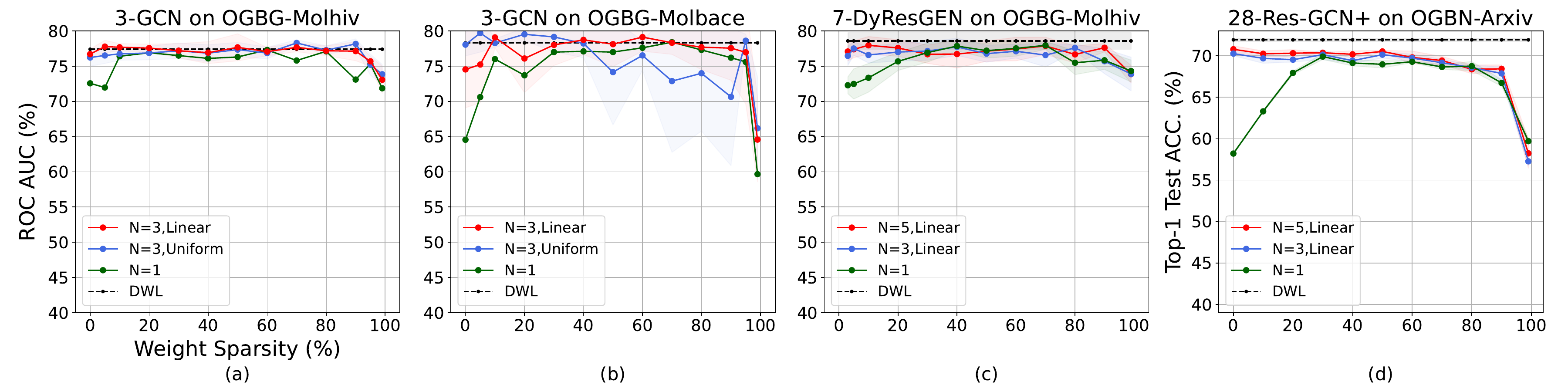}
	\caption{Accuracy of GNN models across various sparsity values with M-Sup ($N = \left \{3, 5\right \}$) and S-Sup ($N = 1$, UGT~\cite{UGT_LOG'22}): (a) and (b) depict shallow GNNs utilizing a 3-layer GCN with a width of 448. (c) and (d) are about deep GNNs, where (c) shows a 7-layer DyResGEN on OGBG-Molhiv, and (d) focuses on a 28-layer ResGCN+ applied to OGBN-Arxiv.} 
	\label{fig:3GCN-graphlevel}
    \vspace*{-0.3cm}  
\end{figure}

The performance of GCN, GAT, and GIN on node-level tasks is illustrated in Figure~\ref{fig:shallow-GCN-nodel-level-tasks}. When sparsity is near 0\%, M-Sup outperforms S-Sup's accuracy. As the sparsity increases from 0\% to 20\%, although the S-Sup approach sees a sharp accuracy increase, it still trails behind M-Sup. M-Sup maintains consistent accuracy without apparent drops, even with high sparsity ($80\sim 90\%$). S-Sup models show better accuracy in the 20\%-75\% sparsity range than other ranges, occasionally even achieving comparable accuracy as DWL models. Based on these GNN models, Figure~\ref{fig:tsne-gnn} (b) visualizes the node representations learned by M-Sup and S-Sup. The projection of node representations learned by M-Sup maintains a distinguishable effect. Moreover, the phenomenon persists when fixing the layer and adjusting the number of hidden neurons, especially at lower sparsity values. The results are detailed in Appendix~\ref{append:increasewithmsup}. GCN's performance on graph-level tasks is shown in Figure~\ref{fig:3GCN-graphlevel} (a) and (b). M-Sup outperforms S-Sup and can achieve accuracy comparable to a dense-weight learning model.



\textbf{Deep GNNs in node- and graph-level tasks.} We evaluate the method on deep GNNs on both node- and graph-level tasks. Specifically, we employ a 28-layer ResGCN+ for the OGBN-Arxiv dataset and a 7-layer DyResGEN for the OGBG-Molhiv dataset. The numbers of coats $N=\left \{ 3,5 \right \} $ with the Linear threshold. As shown in Figure~\ref{fig:3GCN-graphlevel} (c) and (d), M-Sup $N=\left \{ 3,5 \right \} $ surpasses the performance of S-Sup. However, the difference in accuracy between $N=3$ and $N=5$ is not apparent. 

In addition, comparing the Uniform threshold and Linear threshold, Figure~\ref{fig:shallow-GCN-nodel-level-tasks} shows no apparent difference on Cora, Citeseer, and Pubmen. On OGBN-Arxiv with GAT, the Linear threshold outperforms, but the Uniform threshold is better with GCN and GIN. The superior approach potentially depends on the neural networks and datasets being used.

\subsection{Deep GNNs are folded with SLTH}
This subsection explores applying folding methods, including MSF and unshared masks, to deep GNNs. Our results demonstrate that folding methods in combination with SLTH can attain accuracy comparable to the original feed-forward counterparts while enhancing memory efficiency. 

\begin{figure}[tb]
	\centering
	\includegraphics[width=1.0\linewidth]{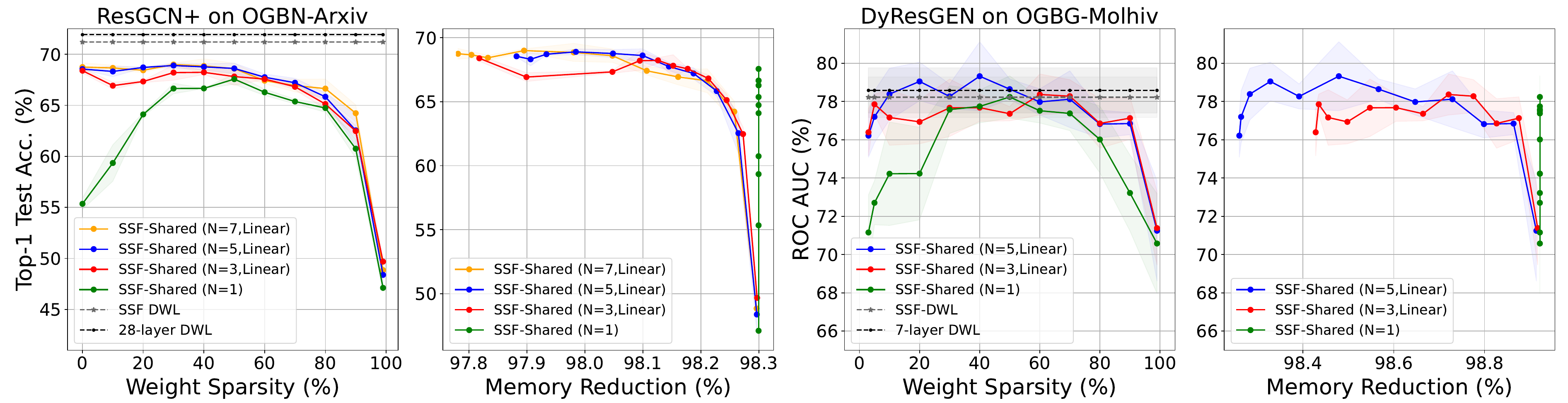}
	\caption{Comparison of accuracy and memory reduction between deep GNNs and DWL models: The figure illustrates using M-Sup ($N={3, 5, 7}$) and S-Sup ($N=1$) for ResGCN+ and DyResGEN. SSF-DWL represents folding the original model into a single stage with dense-weight learning. The number of hidden neurons are 128 for ResGCN+ and 256 for DyResGEN.}
    \vspace*{-0.5cm}  
	\label{fig:resgcn-dyresgen-pf-shared-sm-mm}
\end{figure}

\textbf{Single-stage folding method with shared masks shows comparable accuracy and high memory efficiency.} This study evaluates the single-stage folding (SSF) method with shared masks on deep GNNs and compares their accuracy and memory reduction with the baseline dense-weight leaning (DWL) models. As depicted in Figure~\ref{fig:resgcn-dyresgen-pf-shared-sm-mm}, M-Sup consistently outperforms S-Sup regarding accuracy across all folded models. In M-Sup, ResGCN+ models employing SSF-Shared achieve competitive accuracy on OGBN-Arxiv, with more than $97\%$ memory reduction. DyResGEN models using SSF-Shared surpass baseline models in accuracy on OGBG-Molhiv, with a $98\%$ memory reduction.

\begin{figure}[h]
	\centering
	\includegraphics[width=1.0\linewidth]{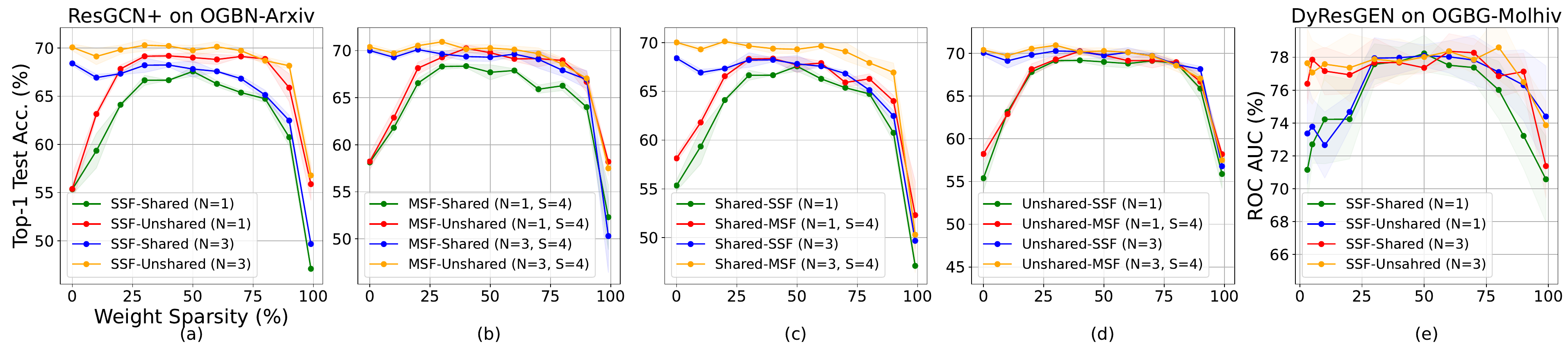}
	\caption{28-layer ResGCN+ and 7-layer DyResGEN with different folding methods on the OGBN-Arxiv and OGBG-Molhiv datasets: (a)-(d) ResGCN+. and (e) DyResGEN. (a) SSF, Shared and Unshared masks; (b) MSF, Shared and Unshared masks; (c) Shared Mask, SSF and MSF; (d) Unshared Mask, SSF and MSF; (e) SSF, Shared and Unshared masks.}
    \vspace*{-0.4cm}  
	\label{fig:Exp-2-Folded-ResGCN}
\end{figure}

\textbf{Unshared masks outperform shared masks with both MSF and SSF methods. Likewise, the MSF method outperforms SSF with both shared and unshared masks.} Furthermore, to evaluate the effectiveness of the unshared and MSF, this work applies them to deep GNNs. Specifically, unshared masks are utilized in ResGCN+ and DyResGEN, while MSF is applied to the 28-layer ResGCN+, which is suitable to be folded into four stages (S = 4). In the case of the ResGCN+, experiments are segmented into four groups, each characterized by different \{folding, sharing\} configurations. Within each group, one of the settings remains constant while the other is altered, accommodating both S-Sup and M-Sup ($N=3$, Linear). For DyResGEN, experiments are segmented into two groups, which have distinct \{sharing\} configurations. A comparative analysis reveals that the unshared mask method outperforms the shared mask method in SSF and MSF, as illustrated in Figure~\ref{fig:Exp-2-Folded-ResGCN}.a, Figure~\ref{fig:Exp-2-Folded-ResGCN}.b and Figure~\ref{fig:Exp-2-Folded-ResGCN}.e. Likewise, in contrast to the SSF method, the MSF method outperforms in shared and unshared mask configurations, as depicted in Figure~\ref{fig:Exp-2-Folded-ResGCN}.c and Figure~\ref{fig:Exp-2-Folded-ResGCN}.d.

Although unshared masks and MSF enhance accuracy, they result in an increase in both memory and parameters. This creates a trade-off which will be discussed in the following subsection.

\subsection{Trade off between the parameter count, memory size, and accuracy}

Optimized GNN models show high memory efficiency with competitive performance. When considering the MSF and unshared mask methods, scores are increased compared to the SSF and shared mask methods. Although the model size increases in MSF and unshared mask methods, the additional parameters are only single-bit, and the accuracy is improved. This subsection explores this trade-off by evaluating optimized GNNs on OGBs.

\begin{figure}[h]
	\centering
	\includegraphics[width=1.0\linewidth]{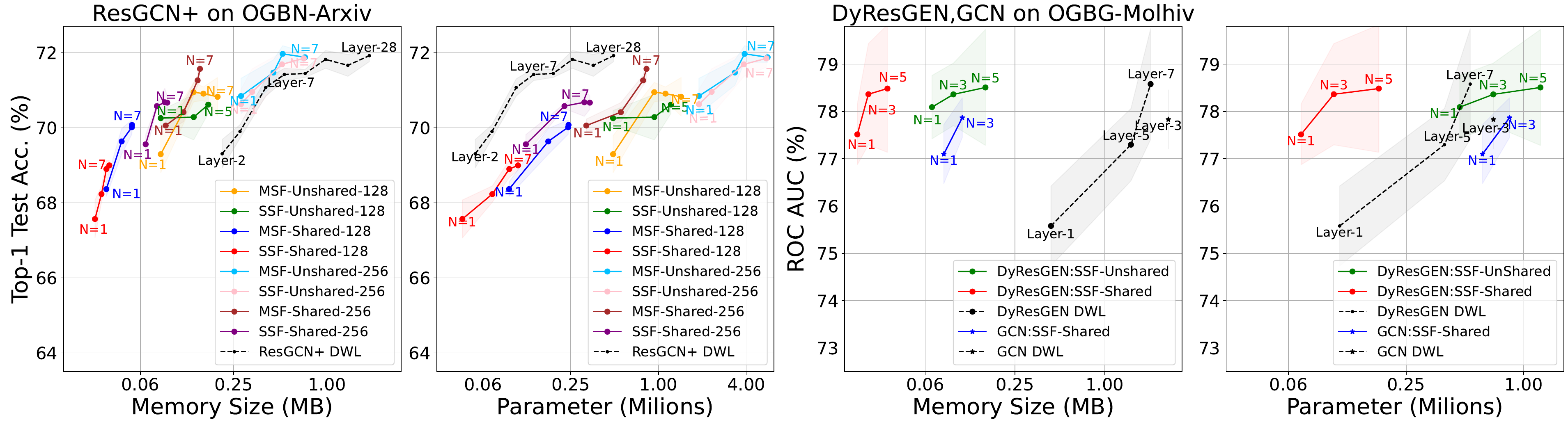}
	\caption{Comparison of the optimized GNN models with different methods on OGBs: optimized ResGCN+ models include $\left \{ 128, 256 \right \}$ hidden neurons, S-Sup ($N=1$, UGT), M-Sup ($N =\left \{ 3, 5, 7 \right \}$, Linear threshold), shared and unshared masks. ResGCN+ DWL are baseline models with dense-weight training; their depths contain $\left \{2, 3, 5, 7, 10, 14, 20, 28 \right \}$. Optimized DyResGEN models contain S-Sup ($N=1$, UGT), M-Sup ($N =\left \{ 3, 5 \right \}$, Linear method), shared and unshared masks. DyResGEN baseline model uses $\left \{1, 5, 7 \right \}$ layers with dense-weight training. Optimized GCN model with three layers contains S-Sup ($N=1$, UGT), M-Sup ($N = 3$, Linear method), and shared masks.}
    \vspace*{-0.4cm}  
	\label{fig:Exp-4TD}
\end{figure}

Figure~\ref{fig:Exp-4TD} compares the accuracy, memory size, and parameter count of various models across different datasets. For DyResGEN in OGBG-Molhiv, SSF-Shared models are the most parameter-efficient and memory-efficient. Compared to a 7-layer DyResGEN, these models achieve a parameter and memory reduction, with 79.9\% fewer parameters and 98.7\% less memory, while maintaining comparable accuracy. Conversely, SSF-Unshared models require more parameters and memory but can achieve slightly better accuracy. The optimized 3-layer GCN on OGBG-Molhive also achieves comparable accuracy to the baseline model with a 95.8\% memory reduction. For ResGCN+ on OGBG-Arixv, this study investigates two hidden neurons, 128 and 256, and utilizes M-Sup with N-values of ${1, 3, 5, 7}$ with SSF and MSF (stage = 4) structures. Supermasks are applied in both shared and unshared manners. Among them, the SSF-Shared-128 models with S-Sup achieve a 77.9\% reduction in parameters and a 97.9\% reduction in memory, albeit with an accuracy 2.8\% lower than the 28-Layer ResGCN model. In contrast, SSF-Unshared-256 and MSF-Unshared-256 models with S-Sup attain a 72\% memory reduction while maintaining accuracy comparable to the 28-Layer ResGCN model. Exact performance is also shown in Appendix~\ref{append:exactdata}. Within this space, various choices emerge, offering multiple candidates for balancing trade-offs between memory, parameters, and performance.



\section{Conclusion}
To enhance the performance of GNNs based on strong lottery tickets, we first applied multicoated supermasks to GNNs with adaptive thresholds. This approach surpasses the previous SLTH method, named UGT. We further reveal the existence of untrained recurrent subnetworks within deep GNNs. By employing Multi-stage Folding and Unshared Mask methods, we explore a larger search space for deep GNNs, achieving significant memory reduction while maintaining an accuracy comparable to the baseline models. These findings are supported by experiments on widely-used GNN models, encompassing diverse datasets, including the Open Graph Benchmark (OGB). This work contributes to energy-efficient graph processing.
\section*{Acknowledgements}
\begin{flushleft}
This work was partially supported by Kakenhi Grant Number 23H05489.
\end{flushleft}
\bibliographystyle{unsrtnat}
\bibliography{reference}

\begin{thebibliography}{44}
\providecommand{\natexlab}[1]{#1}
\providecommand{\url}[1]{\texttt{#1}}
\expandafter\ifx\csname urlstyle\endcsname\relax
  \providecommand{\doi}[1]{doi: #1}\else
  \providecommand{\doi}{doi: \begingroup \urlstyle{rm}\Url}\fi

\bibitem[Kipf and Welling(2017)]{GCN}
Thomas~N. Kipf and Max Welling.
\newblock Semi-supervised classification with graph convolutional networks.
\newblock In \emph{International Conference on Learning Representations}, 2017.

\bibitem[Veličković et~al.(2018)Veličković, Cucurull, Casanova, Romero, Liò, and Bengio]{GAT}
Petar Veličković, Guillem Cucurull, Arantxa Casanova, Adriana Romero, Pietro Liò, and Yoshua Bengio.
\newblock Graph attention networks.
\newblock In \emph{International Conference on Learning Representations}, 2018.

\bibitem[Xu et~al.(2019)Xu, Hu, Leskovec, and Jegelka]{GIN}
Keyulu Xu, Weihua Hu, Jure Leskovec, and Stefanie Jegelka.
\newblock How powerful are graph neural networks?
\newblock In \emph{International Conference on Learning Representations}, 2019.

\bibitem[Li et~al.(2019)Li, Muller, Thabet, and Ghanem]{li2019deepgcns}
Guohao Li, Matthias Muller, Ali Thabet, and Bernard Ghanem.
\newblock Deepgcns: Can {GCN}s go as deep as {CNN}s?
\newblock In \emph{Proceedings of the IEEE/CVF international conference on computer vision}, pages 9267--9276, 2019.

\bibitem[Li et~al.(2020)Li, Xiong, Thabet, and Ghanem]{li2020deepergcn}
Guohao Li, Chenxin Xiong, Ali Thabet, and Bernard Ghanem.
\newblock Deeper{GCN}: All you need to train deeper {GCN}s.
\newblock \emph{arXiv preprint arXiv:2006.07739}, 2020.

\bibitem[Chen et~al.(2021)Chen, Sui, Chen, Zhang, and Wang]{chen2021unified}
Tianlong Chen, Yongduo Sui, Xuxi Chen, Aston Zhang, and Zhangyang Wang.
\newblock A unified lottery ticket hypothesis for graph neural networks.
\newblock In \emph{International conference on machine learning}, pages 1695--1706. PMLR, 2021.

\bibitem[Wang et~al.(2022)Wang, Liang, Wang, Wang, Gu, Fang, and Wang]{wang2022searching}
Kun Wang, Yuxuan Liang, Pengkun Wang, Xu~Wang, Pengfei Gu, Junfeng Fang, and Yang Wang.
\newblock Searching lottery tickets in graph neural networks: A dual perspective.
\newblock In \emph{The Eleventh International Conference on Learning Representations}, 2022.

\bibitem[Hui et~al.(2023)Hui, Yan, Ma, and Ku]{hui2023rethinking}
Bo~Hui, Da~Yan, Xiaolong Ma, and Wei-Shinn Ku.
\newblock Rethinking graph lottery tickets: Graph sparsity matters.
\newblock In \emph{The Eleventh International Conference on Learning Representations}, 2023.

\bibitem[Huang et~al.(2022)Huang, Chen, Fang, Menkovski, Zhao, Yin, Pei, Mocanu, Wang, Pechenizkiy, and Liu]{UGT_LOG'22}
Tianjin Huang, Tianlong Chen, Meng Fang, Vlado Menkovski, Jiaxu Zhao, Lu~Yin, Yulong Pei, Decebal~Constantin Mocanu, Zhangyang Wang, Mykola Pechenizkiy, and Shiwei Liu.
\newblock You can have better graph neural networks by not training weights at all: Finding untrained {GNN}s tickets.
\newblock In \emph{The First Learning on Graphs Conference}, 2022.

\bibitem[Frankle and Carbin(2019)]{lottery_theory}
Jonathan Frankle and Michael Carbin.
\newblock The lottery ticket hypothesis: Finding sparse, trainable neural networks.
\newblock In \emph{7th International Conference on Learning Representations, {ICLR} 2019, New Orleans, LA, USA, May 6-9, 2019}. OpenReview.net, 2019.

\bibitem[Ramanujan et~al.(2020)Ramanujan, Wortsman, Kembhavi, Farhadi, and Rastegari]{edgepopup}
Vivek Ramanujan, Mitchell Wortsman, Aniruddha Kembhavi, Ali Farhadi, and Mohammad Rastegari.
\newblock What's hidden in a randomly weighted neural network?
\newblock In \emph{Proceedings of the IEEE/CVF conference on computer vision and pattern recognition}, pages 11893--11902, 2020.

\bibitem[You et~al.(2019)You, Li, Xu, Fu, Wang, Chen, Baraniuk, Wang, and Lin]{you2019drawing}
Haoran You, Chaojian Li, Pengfei Xu, Yonggan Fu, Yue Wang, Xiaohan Chen, Richard~G Baraniuk, Zhangyang Wang, and Yingyan Lin.
\newblock Drawing early-bird tickets: Toward more efficient training of deep networks.
\newblock In \emph{International Conference on Learning Representations}, 2019.

\bibitem[Lee et~al.(2019)Lee, Ajanthan, Gould, and Torr]{lee2019signal}
Namhoon Lee, Thalaiyasingam Ajanthan, Stephen Gould, and Philip~HS Torr.
\newblock A signal propagation perspective for pruning neural networks at initialization.
\newblock In \emph{International Conference on Learning Representations}, 2019.

\bibitem[Wang et~al.(2020)Wang, Zhang, Xie, Zhou, Su, Zhang, and Hu]{wang2020pruning}
Yulong Wang, Xiaolu Zhang, Lingxi Xie, Jun Zhou, Hang Su, Bo~Zhang, and Xiaolin Hu.
\newblock Pruning from scratch.
\newblock In \emph{Proceedings of the AAAI Conference on Artificial Intelligence}, volume~34, pages 12273--12280, 2020.

\bibitem[Wang et~al.(2019)Wang, Zhang, and Grosse]{wang2019picking}
Chaoqi Wang, Guodong Zhang, and Roger Grosse.
\newblock Picking winning tickets before training by preserving gradient flow.
\newblock In \emph{International Conference on Learning Representations}, 2019.

\bibitem[Frankle et~al.(2020)Frankle, Dziugaite, Roy, and Carbin]{frankle2020pruning}
Jonathan Frankle, Gintare~Karolina Dziugaite, Daniel Roy, and Michael Carbin.
\newblock Pruning neural networks at initialization: Why are we missing the mark?
\newblock In \emph{International Conference on Learning Representations}, 2020.

\bibitem[Sreenivasan et~al.(2022)Sreenivasan, Sohn, Yang, Grinde, Nagle, Wang, Xing, Lee, and Papailiopoulos]{sreenivasan2022rare}
Kartik Sreenivasan, Jy-yong Sohn, Liu Yang, Matthew Grinde, Alliot Nagle, Hongyi Wang, Eric Xing, Kangwook Lee, and Dimitris Papailiopoulos.
\newblock Rare gems: Finding lottery tickets at initialization.
\newblock \emph{Advances in Neural Information Processing Systems}, 35:\penalty0 14529--14540, 2022.

\bibitem[Zhou et~al.(2019)Zhou, Lan, Liu, and Yosinski]{zhou2019deconstructing}
Hattie Zhou, Janice Lan, Rosanne Liu, and Jason Yosinski.
\newblock Deconstructing lottery tickets: Zeros, signs, and the supermask.
\newblock \emph{Advances in neural information processing systems}, 32, 2019.

\bibitem[Malach et~al.(2020)Malach, Yehudai, Shalev-Schwartz, and Shamir]{malach2020proving}
Eran Malach, Gilad Yehudai, Shai Shalev-Schwartz, and Ohad Shamir.
\newblock Proving the lottery ticket hypothesis: Pruning is all you need.
\newblock In \emph{International Conference on Machine Learning}, pages 6682--6691. PMLR, 2020.

\bibitem[Pensia et~al.(2020)Pensia, Rajput, Nagle, Vishwakarma, and Papailiopoulos]{pensia2020optimal}
Ankit Pensia, Shashank Rajput, Alliot Nagle, Harit Vishwakarma, and Dimitris Papailiopoulos.
\newblock Optimal lottery tickets via subset sum: Logarithmic over-parameterization is sufficient.
\newblock \emph{Advances in neural information processing systems}, 33:\penalty0 2599--2610, 2020.

\bibitem[Orseau et~al.(2020)Orseau, Hutter, and Rivasplata]{orseau2020logarithmic}
Laurent Orseau, Marcus Hutter, and Omar Rivasplata.
\newblock Logarithmic pruning is all you need.
\newblock \emph{Advances in Neural Information Processing Systems}, 33:\penalty0 2925--2934, 2020.

\bibitem[Okoshi et~al.(2022)Okoshi, Garc{\'\i}a-Arias, Hirose, Ando, Kawamura, Van~Chu, Motomura, and Yu]{okoshi2022multicoated}
Yasuyuki Okoshi, Ángel~L{\'o}pez Garc{\'\i}a-Arias, Kazutoshi Hirose, Kota Ando, Kazushi Kawamura, Thiem Van~Chu, Masato Motomura, and Jaehoon Yu.
\newblock Multicoated supermasks enhance hidden networks.
\newblock In \emph{Proc. Int. Conf. Mach. Learn.}, pages 17045--17055, 2022.

\bibitem[Yan et~al.(2020)Yan, Deng, Hu, Liang, Feng, Ye, Zhang, Fan, and Xie]{yan2020hygcn}
Mingyu Yan, Lei Deng, Xing Hu, Ling Liang, Yujing Feng, Xiaochun Ye, Zhimin Zhang, Dongrui Fan, and Yuan Xie.
\newblock Hygcn: A gcn accelerator with hybrid architecture.
\newblock In \emph{2020 IEEE International Symposium on High Performance Computer Architecture (HPCA)}, pages 15--29. IEEE, 2020.

\bibitem[Yoo et~al.(2023)Yoo, Song, Lee, Kim, Kim, and Lee]{yoo2023sgcn}
Mingi Yoo, Jaeyong Song, Jounghoo Lee, Namhyung Kim, Youngsok Kim, and Jinho Lee.
\newblock Sgcn: Exploiting compressed-sparse features in deep graph convolutional network accelerators.
\newblock In \emph{2023 IEEE International Symposium on High-Performance Computer Architecture (HPCA)}, pages 1--14. IEEE, 2023.

\bibitem[Hirose et~al.(2022)Hirose, Yu, Ando, Okoshi, Garc{\'\i}a-Arias, Suzuki, Van~Chu, Kawamura, and Motomura]{hirose2022hiddenite}
Kazutoshi Hirose, Jaehoon Yu, Kota Ando, Yasuyuki Okoshi, {\'A}ngel~L{\'o}pez Garc{\'\i}a-Arias, Junnosuke Suzuki, Thiem Van~Chu, Kazushi Kawamura, and Masato Motomura.
\newblock Hiddenite: {4K}-{PE} hidden network inference {4D}-tensor engine exploiting on-chip model construction achieving 34.8-to-16.0 {TOPS/W} for {CIFAR}-100 and {ImageNet}.
\newblock In \emph{2022 IEEE International Solid-State Circuits Conference (ISSCC)}, volume~65, pages 1--3. IEEE, 2022.

\bibitem[Garc{\'{\i}}a{-}Arias et~al.(2023)Garc{\'{\i}}a{-}Arias, Okoshi, Hashimoto, Motomura, and Yu]{lopez'access}
{\'{A}}ngel~L{\'{o}}pez Garc{\'{\i}}a{-}Arias, Yasuyuki Okoshi, Masanori Hashimoto, Masato Motomura, and Jaehoon Yu.
\newblock Recurrent residual networks contain stronger lottery tickets.
\newblock \emph{{IEEE} Access}, 11:\penalty0 16588--16604, 2023.
\newblock \doi{10.1109/ACCESS.2023.3245808}.

\bibitem[Garc{\'{\i}}a{-}Arias et~al.(2021)Garc{\'{\i}}a{-}Arias, Hashimoto, Motomura, and Yu]{Garcia-AriasHMY21}
{\'{A}}ngel~L{\'{o}}pez Garc{\'{\i}}a{-}Arias, Masanori Hashimoto, Masato Motomura, and Jaehoon Yu.
\newblock Hidden-fold networks: Random recurrent residuals using sparse supermasks.
\newblock In \emph{32nd British Machine Vision Conference 2021, {BMVC} 2021, Online, November 22-25, 2021}, page 205. {BMVA} Press, 2021.

\bibitem[Hamilton et~al.(2017)Hamilton, Ying, and Leskovec]{hamilton2017inductive}
Will Hamilton, Zhitao Ying, and Jure Leskovec.
\newblock Inductive representation learning on large graphs.
\newblock \emph{Advances in neural information processing systems}, 30, 2017.

\bibitem[Corso et~al.(2020)Corso, Cavalleri, Beaini, Li{\`o}, and Veli{\v{c}}kovi{\'c}]{corso2020principal}
Gabriele Corso, Luca Cavalleri, Dominique Beaini, Pietro Li{\`o}, and Petar Veli{\v{c}}kovi{\'c}.
\newblock Principal neighbourhood aggregation for graph nets.
\newblock \emph{Advances in Neural Information Processing Systems}, 33:\penalty0 13260--13271, 2020.

\bibitem[Ying(2018)]{ying2018jiaxuan}
Zhitao Ying.
\newblock Jiaxuan you, christopher morris, xiang ren, will hamilton, and jure leskovec. hierarchical graph representation learning with differentiable pooling.
\newblock \emph{Advances in neural information processing systems}, 31:\penalty0 4800--4810, 2018.

\bibitem[Li et~al.(2018)Li, Han, and Wu]{li2018deeper}
Qimai Li, Zhichao Han, and Xiao-Ming Wu.
\newblock Deeper insights into graph convolutional networks for semi-supervised learning.
\newblock In \emph{Proceedings of the AAAI conference on artificial intelligence}, volume~32, 2018.

\bibitem[He et~al.(2016)He, Zhang, Ren, and Sun]{he2016deep}
Kaiming He, Xiangyu Zhang, Shaoqing Ren, and Jian Sun.
\newblock Deep residual learning for image recognition.
\newblock In \emph{Proceedings of the IEEE conference on computer vision and pattern recognition}, pages 770--778, 2016.

\bibitem[Yu and Koltun(2015)]{yu2015multi}
Fisher Yu and Vladlen Koltun.
\newblock Multi-scale context aggregation by dilated convolutions.
\newblock \emph{arXiv preprint arXiv:1511.07122}, 2015.

\bibitem[Rong et~al.(2019)Rong, Huang, Xu, and Huang]{dropedge}
Yu~Rong, Wenbing Huang, Tingyang Xu, and Junzhou Huang.
\newblock Dropedge: Towards deep graph convolutional networks on node classification.
\newblock \emph{arXiv preprint arXiv:1907.10903}, 2019.

\bibitem[Huang et~al.(2020)Huang, Rong, Xu, Sun, and Huang]{dropnode}
Wenbing Huang, Yu~Rong, Tingyang Xu, Fuchun Sun, and Junzhou Huang.
\newblock Tackling over-smoothing for general graph convolutional networks.
\newblock \emph{arXiv preprint arXiv:2008.09864}, 2020.

\bibitem[Zhou et~al.(2021)Zhou, Dong, Wang, Lee, Hooi, Xu, and Feng]{nodenorm}
Kuangqi Zhou, Yanfei Dong, Kaixin Wang, Wee~Sun Lee, Bryan Hooi, Huan Xu, and Jiashi Feng.
\newblock Understanding and resolving performance degradation in deep graph convolutional networks.
\newblock In \emph{Proceedings of the 30th ACM International Conference on Information \& Knowledge Management}, pages 2728--2737, 2021.

\bibitem[Zhao and Akoglu(2019)]{zhao2019pairnorm}
Lingxiao Zhao and Leman Akoglu.
\newblock Pairnorm: Tackling oversmoothing in {GNN}s.
\newblock \emph{arXiv preprint arXiv:1909.12223}, 2019.

\bibitem[Greff et~al.(2016)Greff, Srivastava, and Schmidhuber]{greff2016highway}
Klaus Greff, Rupesh~K Srivastava, and J{\"u}rgen Schmidhuber.
\newblock Highway and residual networks learn unrolled iterative estimation.
\newblock \emph{arXiv preprint arXiv:1612.07771}, 2016.

\bibitem[Liao and Poggio(2016)]{liao2016bridging}
Qianli Liao and Tomaso Poggio.
\newblock Bridging the gaps between residual learning, recurrent neural networks and visual cortex.
\newblock \emph{arXiv preprint arXiv:1604.03640}, 2016.

\bibitem[Hu et~al.(2020)Hu, Fey, Zitnik, Dong, Ren, Liu, Catasta, and Leskovec]{hu2020open}
Weihua Hu, Matthias Fey, Marinka Zitnik, Yuxiao Dong, Hongyu Ren, Bowen Liu, Michele Catasta, and Jure Leskovec.
\newblock Open graph benchmark: Datasets for machine learning on graphs.
\newblock \emph{Advances in neural information processing systems}, 33:\penalty0 22118--22133, 2020.

\bibitem[Stock et~al.(2021)Stock, Fan, Graham, Grave, Gribonval, Jegou, and Joulin]{QAT}
Pierre Stock, Angela Fan, Benjamin Graham, Edouard Grave, R{\'e}mi Gribonval, Herve Jegou, and Armand Joulin.
\newblock Training with quantization noise for extreme model compression.
\newblock In \emph{ICLR 2021-International Conference on Learning Representations}, 2021.

\bibitem[Tailor et~al.(2020)Tailor, Fernandez-Marques, and Lane]{Degree-Quant}
Shyam~Anil Tailor, Javier Fernandez-Marques, and Nicholas~Donald Lane.
\newblock Degree-quant: Quantization-aware training for graph neural networks.
\newblock In \emph{International Conference on Learning Representations}, 2020.

\bibitem[You et~al.(2022)You, Geng, Zhang, Li, and Lin]{GCOD}
Haoran You, Tong Geng, Yongan Zhang, Ang Li, and Yingyan Lin.
\newblock Gcod: Graph convolutional network acceleration via dedicated algorithm and accelerator co-design.
\newblock In \emph{2022 IEEE International Symposium on High-Performance Computer Architecture (HPCA)}, pages 460--474. IEEE, 2022.

\bibitem[Mu and Kim(2021)]{mu202129}
Junjie Mu and Bongjin Kim.
\newblock 29.2 a 21$\times$ 21 dynamic-precision bit-serial computing graph accelerator for solving partial differential equations using finite difference method.
\newblock In \emph{2021 IEEE International Solid-State Circuits Conference (ISSCC)}, volume~64, pages 406--408. IEEE, 2021.

\end{thebibliography}

\appendix
\newpage
\section{Implementation details} \label{append:implementationde}
In this paper, experiments about shallow GNNs, including 2 and 4-layer GCN, GAT, and GIN, are conducted on 1 TESLA V100 GPU (32GB). Experiments about deep GNNs, including 28-layer ResGCN+ and 7-layer DyResGEN, are implemented on 1 NVIDIA RTX A6000 (48GB). Five independent repeated runs are implemented in all the results reported in this paper, except 28-layer ResGCN+, which uses three repeated runs. The average value and standard deviation of accuracy are based on these independent runs. We follow~\cite{GCN, UGT_LOG'22, li2019deepgcns, li2020deepergcn} to train dense GNN models as baselines. For training GNNs with SLTH, the hyper-parameter configurations are summarized in \tablename~\ref{Tab:idshallowGNNs} and \tablename~\ref{Tab:iddeepGNNs}.


\begin{table}[htbp]
\centering
\caption{Hyperparameters for Shallow GNNs with SLTH}
\small 
{\arrayrulewidth=0.4pt
    \begin{tabular}{l|c|c|c|c|c|c}
    \bhline{1.5pt}  
    Dataset & Cora & Citeser & Pubmed & OGBN-Arxiv & OGBG-Molhiv & OGBG-Molbace \\ \bhline{1.5pt}  
    GNN & \multicolumn{4}{|c|}{GCN, GAT, GIN} & \multicolumn{2}{|c}{GCN}\\ \hline
    Depth & 2 & 2 & 2 & 4 & 3 & 3 \\
    Epochs & 400 & 400 & 400 & 400 & 200 & 200 \\
    Learning Rate & 0.01 & 0.01 & 0.01 & 0.01 & 0.04 & 0.015 \\
    \#Hidden neurons & 256 & 256 & 256 & 386 & 448 & 448 \\
    Optimizer & Adam & Adam & Adam & Adam & Adam & Adam \\
    Weight decay & 0.0 & 0.0 & 0.0 & 0.0 & 0.0 & 0.0 \\
    Batch size & - & - & - & - & 512 & 64 \\ \bhline{1.5pt}  
    \end{tabular}  
}
\label{Tab:idshallowGNNs}
\end{table}

\begin{table}[htbp]
\centering
\caption{Hyperparameters for Deep GNNs with SLTH}
{\arrayrulewidth=0.4pt
    \begin{tabular}{l|c|c}
    \bhline{1.5pt}  
    Dataset & OGBN-Arxiv & OGBG-Molhiv \\ \bhline{1.5pt}  
    GNN & ResGCN+ & DyResGEN \\ \hline
    Depth & SSF, MSF (stage = 4) & SSF \\
    Epochs & 500 & 300 \\
    Learning Rate & 0.01 & 0.016 \\
    \#Hidden neurons & 128, 256 & 256 \\
    Optimizer & Adam & Adam \\
    Weight decay & 0.001 & 0.0 \\
    Batch size & - & 512 \\ \bhline{1.5pt}  
    \end{tabular}  
}
\label{Tab:iddeepGNNs}
\end{table}

For the train-val-test split of the datasets, we follow~\cite{UGT_LOG'22} by using 140 (Cora), 120 (Citeseer), and 60 (PubMed) labeled data for training, 500 nodes for validation and 1,000 nodes for testing. We follow ~\cite{hu2020open} for splitting OGBN-Arxiv dataset. For OGBG-Molhiv, we use 32,901 graphs for training, 4,113 for validation, and 4,113 for testing. For OGBG-Molbace, we use 1,210 graphs for training, 151 for validation, and 152 for testing.

Compared with other pruning methods such as GLT~\cite{hui2023rethinking} and HGS~\cite{wang2022searching}, our proposal does not need to do the fine-tuning or retraining of weights post-pruning, which reduces training time. Specifically, the HGS demands an extra 200 epochs for fine-tuning a shallow GCN on datasets including Cora, Citeseer, and Pubmed after its regular training of 500 epochs. 

\section{Initilization methods for weights} \label{append:SCKN}
\begin{figure}[h!]
	\centering
	\includegraphics[width=1.0\linewidth]{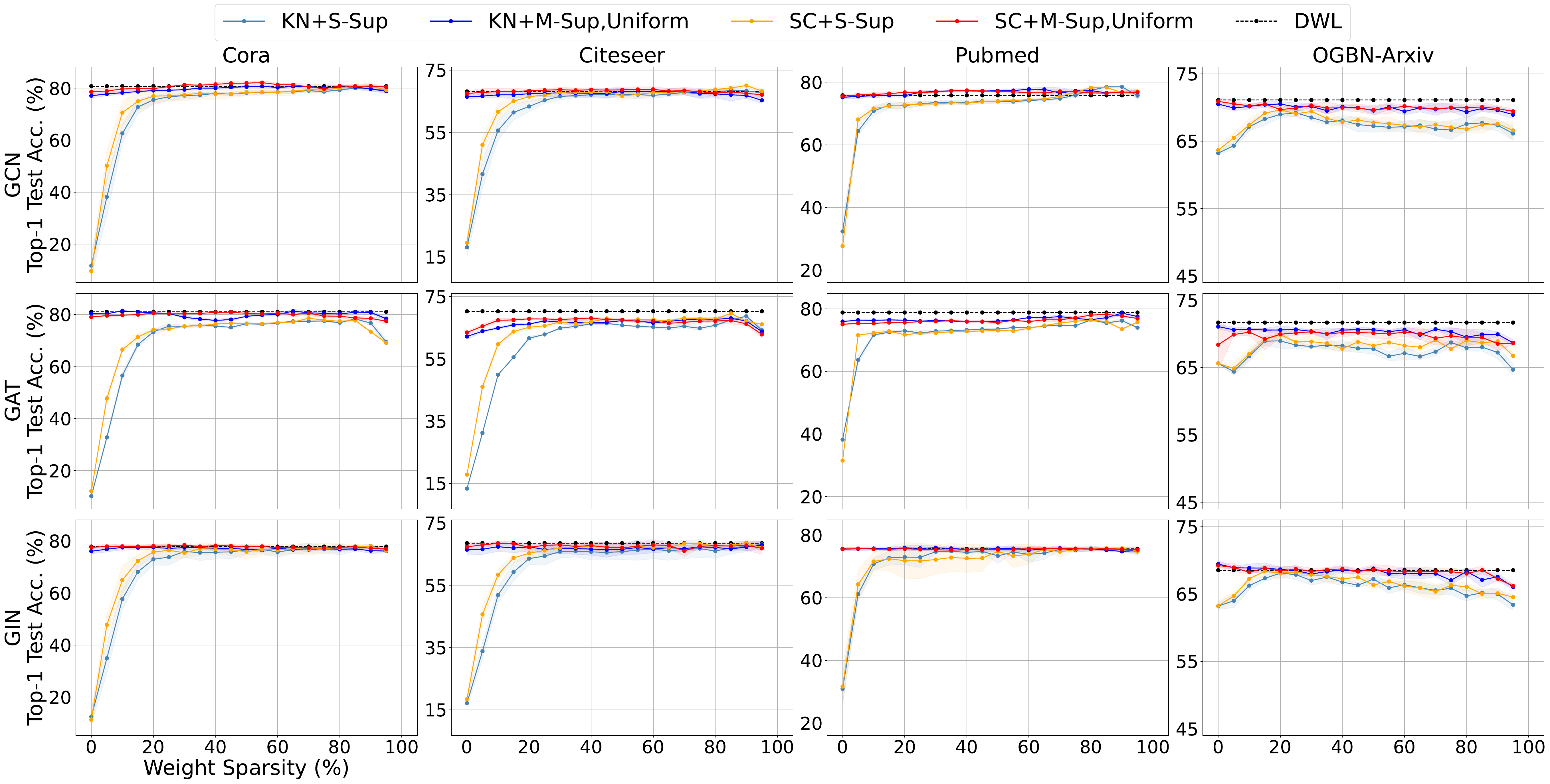}
	\caption{The accuracy of GCN, GAT, and GIN on Cora, Citeseer, Pubmed, and OGBN-Arxiv datasets with SC and KN initialization methods: the models are 256-2-GNN for Cora, Citeseer, and Pubmed, and 386-4-GNN for OGBN-Arxiv with affine batch normalization.}
	\label{fig:inimethods}
\end{figure}

Figure~\ref{fig:inimethods} shows the accuracy of GCN, GAT, and GIN on four datasets with different initialization methods. As the figure shows, Singed Kaiming Constant (SC) outperforms Kaiming Normal (KN) when GNNs are applied with S-Sup, and the SC is slightly better than KN when GNNs are used with M-Sup. 

\section{The adaptive Linear threshold and non-adaptive Linear threshold} \label{append:adaptive}
\begin{figure}[h!]
	\centering
	\includegraphics[width=0.75\linewidth]{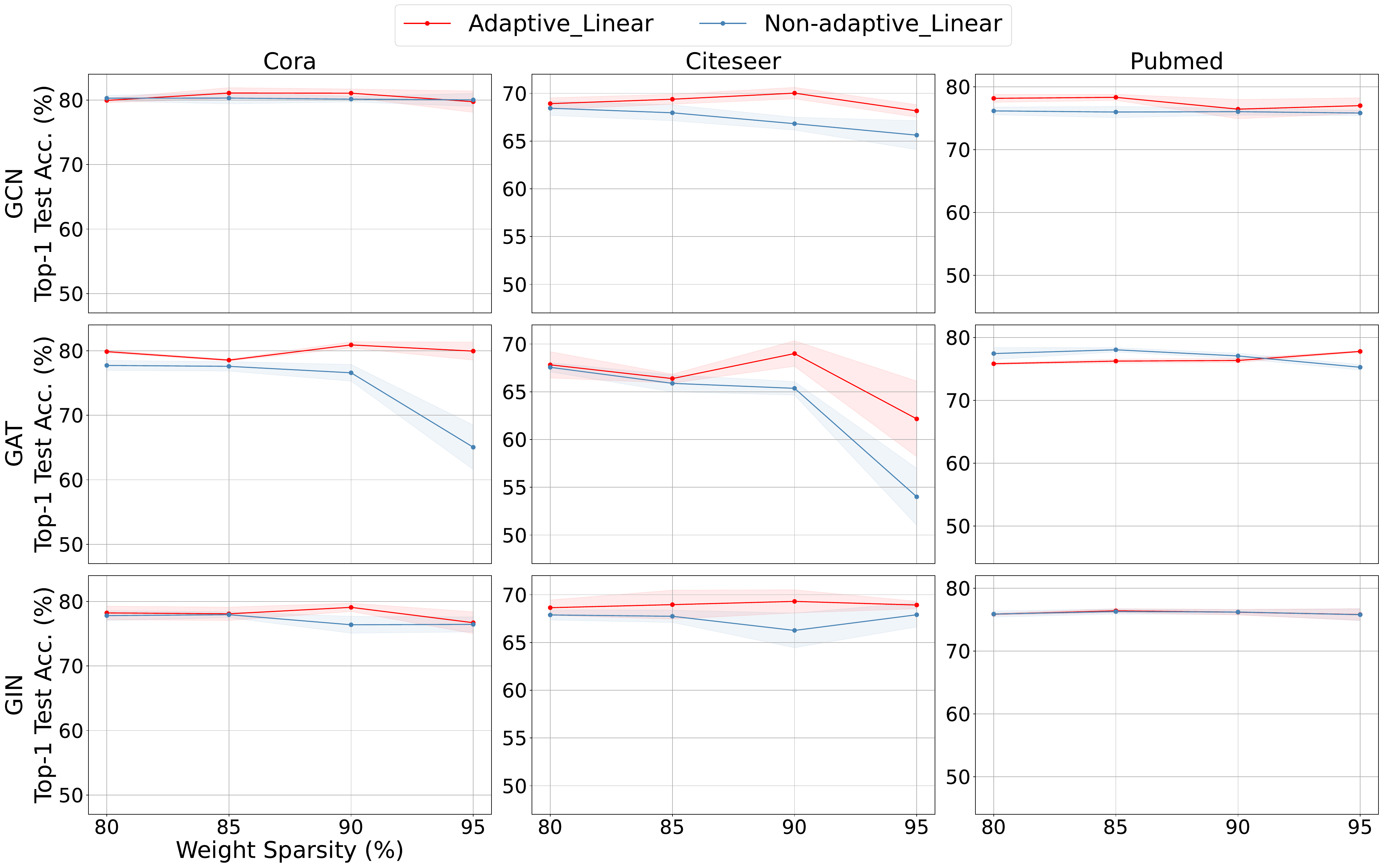}
	\caption{The accuracy of GCN, GAT, and GIN on Cora, Citeseer, Pubmed, and OGBN-Arxiv datasets with different Linear thresholds: the red line is for the adaptive threshold, and the blue line is for the non-adaptive threshold.}
	\label{fig:linear}
\end{figure}

Experiments evaluate  GCN, GAT, and GIN on Cora, Citeseer, and Pubmed datasets with the adaptive Linear threshold ($\alpha=0.9996$) and the non-adaptive Linear threshold ($\alpha=1.0$). In Figure~\ref{fig:linear}, results show when the sparsity becomes high, the adaptive Linear threshold outperforms.

\section{Increasing hidden neurons with S-Sup and M-Sup} \label{append:increasewithmsup}
\begin{figure}[h!]
	\centering
	\includegraphics[width=1.0\linewidth]{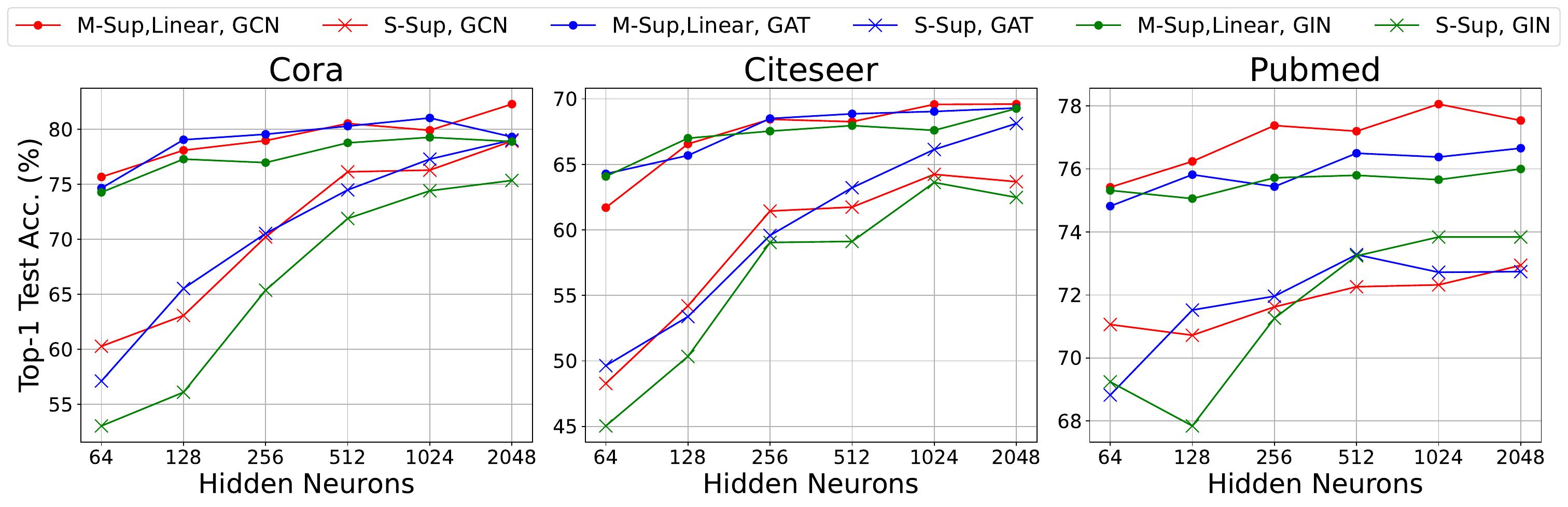}
	\caption{The accuracy of GCN, GAT, and GIN on Cora, Citeseer, and Pubmed datasets with different hidden neurons: experiments set different hidden dimension sizes in 2-layer structures, and the sparsity value is 10\%.}
	\label{fig:increaseneurons}
\end{figure}

Experiments evaluate GCN, GAT, and GIN on Cora, Citeseer, and Pubmed datasets with different hidden neurons. The results are shown in Figure~\ref{fig:increaseneurons}. As the number of hidden neurons increases, the accuracy improves. In addition, the M-Sup method still outperforms better than S-Sup.

\section{Pseudocode for Training} \label{append:pseudocode}
The pseudocode about training weight scores is shown in Algorithm~\ref{alg:fortraining}.

\begin{algorithm}[h!]
	\renewcommand{\algorithmicrequire}{\textbf{Input:}}
	\renewcommand{\algorithmicensure}{\textbf{Output:}}
	\caption{Multicoated and Folded GNN with SLTH} 
	\label{alg:fortraining}
	\begin{algorithmic}[1]
		\STATE \textbf{Input}: a GNN $\mathcal{G\left ( \hat{\mathit{A}}, \mathit{X}, \mathit{W}_\mathrm{rand}\right ) } $, learning rate $\lambda$, hyperparameters for M-Sup and Folding networks: sparsity list $\mathcal{K}=\left \{ k_1,...,k_N \right \} $ with scores $\mathcal{S}$, $m$-stages, $\mathrm{flag_{MSF}}$, $\mathrm{flag_{shared}}$.
        \STATE \textbf{Output}:  $\mathcal{G\left ( \hat{\mathit{A}}, \mathit{X}, \mathit{W}_\mathrm{rand} \odot \sum_{\mathit{k}\in {\mathcal{K}}}^{} \mathcal{H}\left (\mathcal{S},\mathit{k}\right ) \right )},$ $y$
        \STATE Randomly initialize weights $\mathit{W}_\mathrm{rand}$ and Supermasks $\mathcal{S}$ based on $m$-stages, $\mathrm{flag_{MSF}}$, $\mathrm{flag_{shared}}$ as Algorithm~\ref{alg:hidden-folded}.
        \FOR{Iter. $t=1$ to $T$}
            \STATE \textcolor{gray}{\#Get the current sparsity list based on the current epoch, Linear decay schedule as UGT~\cite{UGT_LOG'22}}
            \IF{$t<0.5T$}
                \STATE $\mathcal{K}_{t} = \mathcal{K} \cdot \frac{2t}{T}$  
            \ELSE
                \STATE $\mathcal{K}_{t} = \mathcal{K} $  
            \ENDIF
            \STATE \textcolor{gray}{\#Use the global threshold value based on current $\mathcal{K}_{t}$, and generate the binary masks for weights} 
            \vspace*{-0.3cm}  
            \STATE  $\mathit{W}{ =  \mathit{W}_{\mathrm{rand}} \odot \sum_{k\in \mathcal{K}_{t} }^{} \mathcal{H}(\mathcal{S},k)}$ 
            \STATE \textcolor{gray}{\#Update scores $\mathcal{S}$}
            \STATE $\mathcal{S} = S - \lambda \bigtriangledown_{\mathcal{S}}\mathcal{L}  \left ( \mathcal{G\left ( \hat{\mathit{A}}, \mathit{X}, \mathit{W}_\mathrm{rand} \odot \sum_{\mathit{k}\in {\mathcal{K}_\mathrm{t}}}^{} \mathcal{H}\left (\mathcal{S},\mathit{k}\right ) \right )} \right ) $
        \ENDFOR
        \STATE Return $\mathcal{G\left ( \hat{\mathit{A}}, \mathit{X}, \mathit{W}_\mathrm{rand} \odot \sum_{\mathit{k}\in {\mathcal{K}}}^{} \mathcal{H}\left (\mathcal{S},\mathit{k}\right ) \right )},$ $y$
	\end{algorithmic}  
\end{algorithm}

\section{Small and large sparsity levels for shallow and deep GNNs} \label{append:lhdeepgnns}
To explore the effectiveness in regions of small and large sparsity levels for GNNs, we sampled 1\%, 3\%, 5\%, 7\%, and 9\% data points at the low sparsity level and 91\%, 93\%, 95\%, 97\%, and 99\% at the high sparsity level. The \tablename~\ref{tab:lhdeepgnn} shows shallow and deep GNNs' performance with M-Sup and S-Sup methods. Moreover, single-stage folding (SSF), multi-stage folding (MSF), Shared supermasks, and Unshared supermasks are also explored. We found that the S-Sup method shows a noticeable accuracy deviation at low sparsity levels compared to the dense weight learning model. For instance, ResGCN+ (SSF, Shared) using the S-Sup method has an accuracy gap of roughly 16\%. In contrast, the M-Sup method improves accuracy by 13\% when sparsity is close to 1\%. Utilizing the MSF method with both S-Sup and M-Sup methods increases accuracy, with improvements ranging from 1\% to 4\% for ResGCN+. At high sparsity levels (above 90\%), the M-Sup and S-Sup methods see marked reductions in accuracy.

\begin{table}[h]
\caption{Accuracy (\%) of models in low (< 10\%) and high (> 90\%) sparsity levels}
\centering
\resizebox{\textwidth}{!}{%
\begin{tabular}{c|c|c|c|c|c|c|c|c|c|c|c}
\bhline{1.5pt} 
Models &  Weight Sparsity(\%)  & 1 & 3 & 5 & 7 & 9 & 91 & 93 & 95 & 97 & 99 \\
\bhline{1.5pt} 
{ResGCN+, OGBN-Arxiv} & S-Sup & 55.89  & 56.80  & 55.85  & 57.87 & 58.27  & 61.87  & 58.13  & 57.47  & 52.91 & 47.11  \\
\cline{2-12}
 SSF, Shared & M-Sup & 67.38  & 67.34  & 66.64  & 66.57  & 66.83  & 62.32  & 61.22 & 57.81  & 56.65  & 48.82  \\
\hline
{ResGCN+, OGBN-Arxiv} & S-Sup & 58.26  & 57.61 & 59.57  & 59.19  & 60.81  & 63.22  & 61.93  & 62.17  & 57.71 & 51.52 \\
\cline{2-12}
MSF(4), Shared & M-Sup & 69.54  & 68.62  & 67.73  & 68.75  & 67.88  & 64.30  & 62.30  & 61.13  & 57.64 & 52.24 \\
\hline
DyResGEN, OGBG-Molhiv  & S-Sup & 74.41  & 71.15  & 72.70  & 73.27  & 71.90 & 74.10  & 75.51  & 72.39  & 72.22  & 70.58  \\
\cline{2-12}
SSF, Shared & M-Sup & 78.23  & 77.19  & 77.16  & 76.64  & 77.22  & 77.24 & 75.68 & 75.19  & 73.41 & 70.93  \\
\hline
DyResGEN, OGBG-Molhiv  & S-Sup & 73.10  & 72.47  & 71.36  & 72.57  & 73.26 & 77.26  & 76.51  & 74.36  & 75.48  & 74.05  \\
\cline{2-12}
SSF, Unshared & M-Sup & 79.23  & 78.26  & 78.71  & 75.90  & 77.79  & 76.77 & 75.33 & 76.86  & 76.49 & 74.32  \\
\hline
3-Layer GCN, OGBG-Molhiv, & S-Sup & 71.76 & 70.47 & 69.01 & 69.20 & 69.64 & 75.43 & 75.20 & 74.90 & 74.54 & 70.86 \\
\cline{2-12}
Shared & M-Sup & 74.76 & 76.66 & 74.05 & 75.81 & 74.98 & 75.69 & 75.90 & 75.14 & 74.34 & 72.74 \\
\hline
3-Layer GCN, OGBG-Molbace, & S-Sup & 76.24 & 75.80 & 75.28 & 75.79 & 74.70 & 77.40 & 77.34 & 75.44 & 76.08 & 73.60 \\
\cline{2-12}
Shared & M-Sup & 78.33 & 78.78 & 77.85 & 78.51 & 80.52 & 79.17 & 78.15 & 79.86 & 73.07 & 77.76 \\
\bhline{1.5pt} 
\end{tabular}}
\label{tab:lhdeepgnn}
\begin{minipage}{\textwidth}  
\small
Note: Here are the accuracy of dense-weight learning models. ResGCN+ on OGBN-Arxiv: 71.92±0.16\%, DyResGEN on OGBG-Molhiv: 78.58±1.17\%, 3-layer GCN on OGBN-Molhiv: 77.4±1.24\%, 3-layer GCN on OGBG-Molbace: 78.3±1.59\%.
\end{minipage}
\end{table}

\section{More datasets for DeepGNNs } \label{append:moredatasets}
In addition to OGBN-Arxiv and OGBG-Molhiv, we assessed our proposed methods on three more datasets: Cora, Pubmed, and OGBG-Molbace. Five independent repeated runs are implemented. The OGBG-Molbace dataset is about a graph-level task, and the Cora and Pubmed datasets are about node-level tasks. We investigated proposed deepGNN techniques, including M-Sup, Single-Stage Folding (SSF), Multi-Stage Folding (MSF), and both shared and unshared supermasks. 

\begin{figure}[h!]
	\centering
	\includegraphics[width=1.0\linewidth]{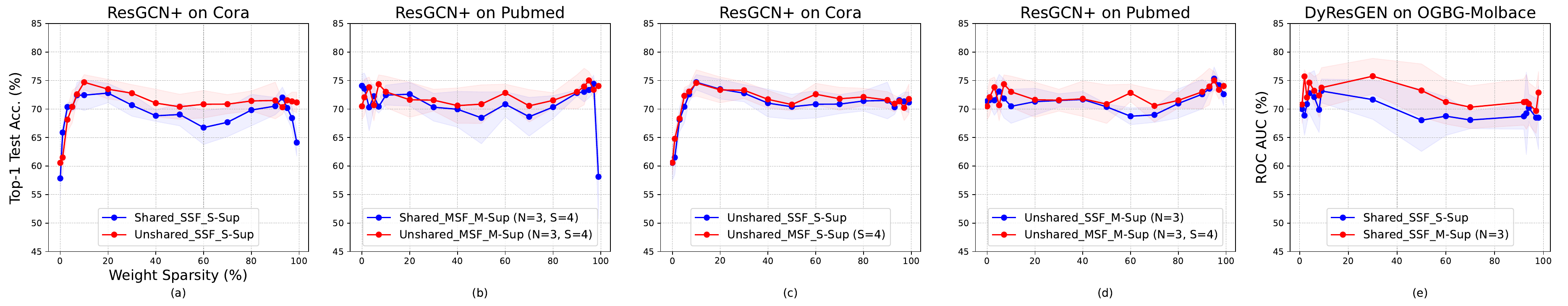}
	\caption{The accuracy of ResGCN+ and DyResGEN on Cora, Pubmed, and OGBG-Molbace datasets: (a) ResGCN+ on Cora with shared and unshared supermasks. (b) ResGCN+ on Pubmed with shared and unshared supermasks. (c) ResGCN+ on Cora dataset with SSF and MSF methods. (d) ResGCN+ on Pubmed with SSF and MSF methods. (e) DyResGEN on OGBG-Molbace with S-Sup and M-Sup methods.}
	\label{fig:moredatasetdeepgnns}
\end{figure}

Figures ~\ref{fig:moredatasetdeepgnns} (a) and (b) compare Shared and Unshared supermasks for the Cora and Pubmed datasets, respectively. The results suggest that using Unshared supermasks can enhance accuracy. Figures ~\ref{fig:moredatasetdeepgnns} (c) and (d) contrast SSF with MSF for the same datasets, indicating a performance boost when employing MSF. Lastly, Figure ~\ref{fig:moredatasetdeepgnns} (e) compares S-Sup and M-Sup on the OGBG-Molbace dataset, revealing better performance of M-Sup over S-Sup.

\section{Exact performance about optimized SLT-GNN models} \label{append:exactdata}
The \tablename~\ref{tab:exactdata} presents the performance of both shallow and deep GNNs across various datasets. For GNNs employing SLT, M-Sup consistently surpasses S-Sup in accuracy. This is particularly noticeable for shallow GNNs, which achieve accuracy levels similar to DWL models. When integrated with unshared supermasks and the multi-stage folding (MSF) technique, the deep GNNs can also rival the accuracy of DWL models. In terms of memory size, our optimized models boast significant memory reductions. For instance, MSF-Unshared-256 achieves a 72\% memory reduction while retaining comparable accuracy. The folding method also contributes to parameter reductions, with the SSF-Shared DyResGEN model achieving a 77.9\% reduction. For inference MACs (multiply–accumulate operations), we focus on the linear layer components as they are optimized using SLT. The reduction in MACs corresponds to the sparsity level: for example, GIN can be reduced by 90\% and ResGCN+ by 40\%.
 
\begin{table}[h]
\caption{Exact performance about optimized SLT-GNN models}
\centering
\resizebox{\textwidth}{!}{%
\begin{tabular}{c|c|c|c|c|c}
\bhline{1.5pt} 
Models &  Configurations & Top-1 Test Acc. / ROC AUC (\%) & Memory Size (MB) & Params (Mils) & MACs (Bils) \\
\bhline{1.5pt}
\multirow{7}{*}{ResGCN+ on OGBN-Arxiv} & 28-layer DWL & 71.92±0.16 & 1.8737 & 0.4912 & 717 \\
\cline{2-6}
                                       & Unfolded-Shared-128 (S-Sup, sparsity=60\%) & 69.28±0.11 ($\downarrow2.64$) & 0.0846 & 0.4874 & 287 \\
\cline{2-6}
                                       & Unfolded-Shared-128 (M-Sup, N=5, sparsity=50\%) & 70.52±0.12 ($\downarrow1.40$) & 0.1329 & 0.8929 & 359 \\
\cline{2-6}
                                       & SSF-Shared-128 (M-Sup, N=5, sparsity=30\%) & 68.89±0.22 ($\downarrow3.03$) & 0.0378 & 0.0946 & 502 \\
\cline{2-6}
                                   & MSF-Shared-128 (M-Sup, N=5, sparsity=10\%) & 69.73±0.25 ($\downarrow2.19$) & 0.0546 & 0.5200 & 646 \\
\cline{2-6} 
(for MACs: full batches)             & \textbf{SSF-Unshared-256 (M-Sup, N=5, sparsity=30\%)} & 71.62±0.21 (\ding{51})  & 0.5678 & 4.3187 & 502 \\
\cline{2-6}
                                       & \textbf{MSF-Unshared-256 (M-Sup, N=5, sparsity=40\%)} & 71.97±0.21 (\ding{51}) & 0.5174 & 3.8960 & 430 \\
\hline 
\multirow{4}{*}{DyResGEN on Molhiv} & 7-layer DWL & 78.58±1.17 & 2.0293 & 0.532 & 3.590 \\
\cline{2-6}
                                      & Unfolded (S-Sup) & 77.20±1.04 ($\downarrow1.38$)& 0.0693 & 0.4703 & 1.795 \\
\cline{2-6}
                                     & \textbf{SSF-Shared (N=3, sparsity=60\%)} & 78.36±1.06 (\ding{51})& 0.0259 & 0.1066 & 1.436 \\
\cline{2-6}
(for MACs: batch = 10 graphs)        & \textbf{SSF-Unshared (N=3, sparsity=60\%)} & 78.36±0.66 (\ding{51})& 0.0964 & 0.6984 & 1.436 \\
\hline 
\multirow{3}{*}{Shallow GCN on Cora} & 3-layer DWL & 80.8±0.71 &  1.4060 & 0.3686 & 0.998 \\
\cline{2-6}
                                     & S-Sup (sparsity=55\%) & 78.5±0.64 ($\downarrow2.30$) &  0.0439 & 0.3686 & 0.449 \\
\cline{2-6}
(for MACs: full batches)      & \textbf{M-Sup (sparsity=55\%)} & 82.1±0.59 (\ding{51}) &  0.0737 & 0.6185 & 0.449 \\
\hline
\multirow{3}{*}{Shallow GAT on Cora} & 3-layer DWL & 81.1±0.34 & 1.4080 & 0.3692 & 0.999 \\
\cline{2-6}
                                     & S-Sup (sparsity=30\%) & 75.6±0.41 ($\downarrow5.50$) &  0.0440 & 0.3692 & 0.699 \\
\cline{2-6}
(for MACs: full batches)      & \textbf{M-Sup (sparsity=30\%)} & 81.7±0.55 (\ding{51}) &  0.0913 & 0.7654 & 0.699 \\
\hline
\multirow{3}{*}{Shallow GIN on Cora} & 3-layer DWL & 77.9±0.72 & 1.4131 & 0.3815 & 1.176 \\
\cline{2-6}
                                     & S-Sup (sparsity=90\%) & 78.0±0.75 (\ding{51}) &  0.0455 & 0.3815 & 0.118 \\
\cline{2-6}
(for MACs: full batches)      &\textbf{ M-Sup (sparsity=90\%)} & 79.1±0.62 (\ding{51}) &  0.0500 & 0.4197 & 0.118 \\
\bhline{1.5pt}
\end{tabular}}
\label{tab:exactdata}
\end{table}

\section{Discussion the differences between proposed methods and other quantization methods} \label{append:quanme}
Considering we only use un-quantized DWL models and S-Sup models as baselines, we further discuss the differences between our methods and quantization techniques. Quantization techniques can be characterized by the choices between asymmetric and symmetric quantization for clipping range, as well as the decision to use static or dynamic scaling values during iterations. We select typical studies in the field of quantization in GNNs, including QAT~\cite{QAT}, Degree-Quant~\cite{Degree-Quant}, and GCoD~\cite{GCOD}. The \tablename~\ref{Tab:quangnns} presents two typical shallow GNNs under different quantization methods. Note that for quantization, both weights and activations are quantized into 8-bit fixed points. Compared with these techniques, our method distinguishes them in three perspectives:   
\begin{enumerate}

\item Our approach integrates both quantization and pruning, with an exclusive emphasis on weights. Rather than designing for hardware's fixed-point processing elements like other quantization methods, we focus on exploring the intrinsic behavior of GNNs, particularly under the strong lottery ticket hypothesis. 
\item Our strategy is bit efficient, employing 2 bits for M-Sup (3 supermasks) and 1 bit for S-Sup (a single supermask). This efficiency translates to notable memory savings compared with 8-bit quantization methods. Using bit-serial multipliers~\cite{mu202129}, our method necessitates only two cycles for multiplication, while an 8-bit approach requires eight cycles. This leads to an empirical improvement in inference time by a factor of four.
\item Regarding accuracy, our method (M-Sup) achieves the comparable accuracy of the FP32 dense-weight learning models in most cases, while for quantization methods, there is still an accuracy gap lower than dense-weight learning models. In the worst-case scenario, QAT on CiteSeer displayed a 4.5\% drop.

\end{enumerate}

\begin{table}
    \centering
    \caption{Comparison of different quantization methods.}
    \resizebox{\textwidth}{!}{%
    \begin{tabular}{c|c|c|c|c|c}
    \bhline{1.5pt} 
    \textbf{Models} & \textbf{Methods} & \textbf{Acc. on Cora (\%)} & \textbf{Acc. on CiteSeer (\%)} & \textbf{Acc. on Pubmed (\%)} & Memory Reduction \\
    \bhline{1.5pt}
    \multirow{4}{*}{GAT (8 heads, 2 layers)} & Vanilla & 83.1 & 72.2 & 78.8 & - \\
    \cline{2-6}
    & QAT~\cite{QAT} & 81.9$(\downarrow1.2)$ & 71.2$(\downarrow1.0)$ & 78.3$(\downarrow0.5)$ \\
    \cline{2-5}
    & Degree-Quant~\cite{Degree-Quant} & 82.7$(\downarrow0.4)$ & 71.6$(\downarrow0.6)$ & 78.6$(\downarrow0.2)$& 75\% with 8-bit Activation\& Weight \\
    \cline{2-5}
    & GCoD~\cite{GCOD} & 82.6$(\downarrow0.5)$ & 71.8$(\downarrow0.4)$ & 78.8 \\
    \hline
    \multirow{4}{*}{GIN (3 layers)} & Vanilla & 78.6 & 67.5 & 78.5 \\
    \cline{2-5}
    & QAT~\cite{QAT} & 75.6$(\downarrow3.0)$ & 63.0$(\downarrow4.5)$ & 77.5$(\downarrow1.0)$ \\
    \cline{2-5}
    & Degree-Quant~\cite{Degree-Quant} & 78.7 & 67.5 & 78.1$(\downarrow0.4)$ & 75\% with 8-bit Activation\& Weight \\
    \cline{2-5}
    & GCoD~\cite{GCOD} & 78.4$(\downarrow0.2)$ & 68.7 & 78.3$(\downarrow0.2)$ \\
    \hline
    \multirow{3}{*}{GAT (1 head, 2 layers)} & Vanilla & 81.1 & 70.3 & 78.8 & - \\
    \cline{2-6}
    & S-Sup & 75.6 $(\downarrow5.50)$ & 67.1 $(\downarrow3.2)$ & 75.8 $(\downarrow3.0)$ & 96.8\% on Cora, 91.9\% on Citeseer, 98.9\% on Pubmed \\
    \cline{2-6}
    & M-Sup& 81.7 ( \ding{51}) & 70.3 (\ding{51}) & 78.4 $(\downarrow0.4)$ & 93.5\% on Cora, 83.3\% on Citeseer, 97.7\% on Pubmed \\
    \hline
    \multirow{3}{*}{GIN (2 layers)} & Vanilla & 77.9   & 68.5 & 75.6 & - \\
    \cline{2-6}
    & S-Sup & 78.0     & 68.4$(\downarrow0.1)$ & 75.7 &  96.7\% on Cora, 91.8\% on Citeseer, 98.1\% on Pubmed\\
    \cline{2-6}
    & M-Sup  & 79.1 ( \ding{51})   & 69.4 ( \ding{51}) & 76.4( \ding{51}) &  96.4\% on Cora, 91.1\% on Citeseer, 98.0\% on Pubmed\\
    \bhline{1.5pt}
    \end{tabular}}
\label{Tab:quangnns}
\end{table}
\begin{figure}[t!]
	\centering
	\includegraphics[width=0.85\linewidth]{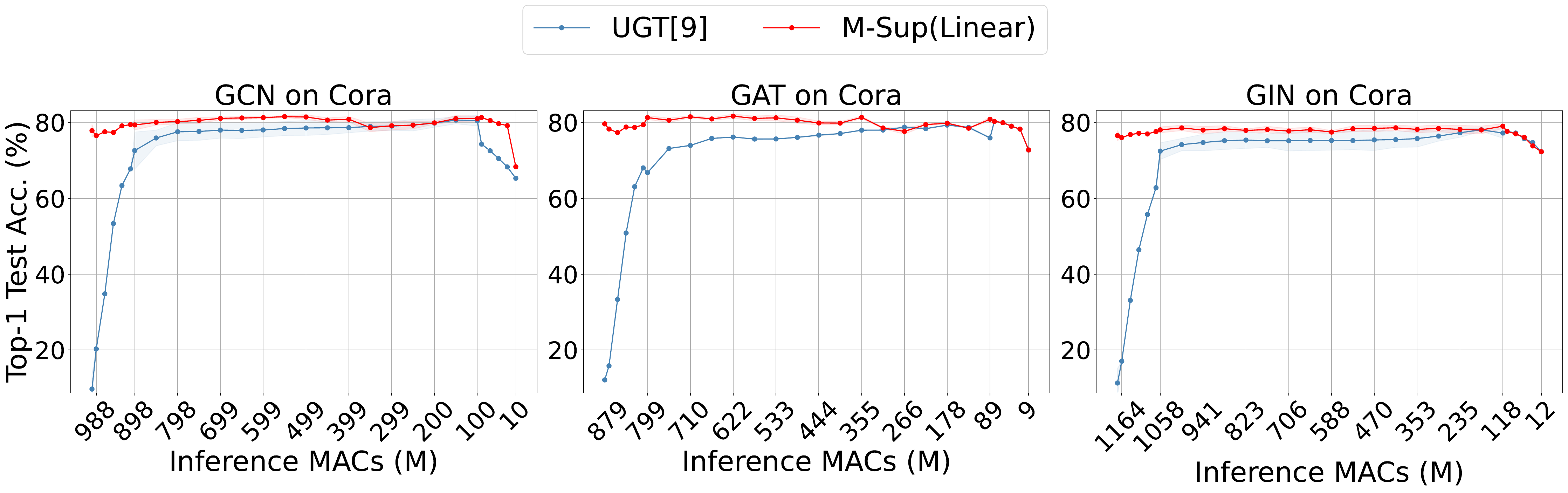}
	\caption{Performance over inference MACs of GCN, GAT, and GIN on Cora.}
	\label{fig:complexity}
\end{figure}

\section{Complexity analysis of our proposal and other pruning methods} \label{append:complexity analysis}
GNN pruning can be performed on both the adjacency matrix and the weight matrix, such as ULTH-GNN~\cite{chen2021unified} and DGLT~\cite{wang2022searching}. In this work, we focus on the weight matrix as well as UGT~\cite{UGT_LOG'22}. The inference time complexity for these methods is given by $\mathcal{O}(||A_{all}||_{0} \times F + ||m_w \odot W ||_0 \times \mathcal{|V|} \times F^2 )$, where $||A_{all}||_{0}$ denotes the number of edges, $F$ represents the dimension of the feature, $\mathcal{|V|}$ is the number of nodes, and $||m_w \odot W ||_0$ signifies the remaining neurons with supermasks. However, our method could also be extended to pruning adjacency matrices like ULTH-GNN~\cite{chen2021unified} and DGLT~\cite{wang2022searching}. A comparative analysis of our proposal with UGT~\cite{UGT_LOG'22} is presented in Figure~\ref{fig:complexity}, where inference MACs are about optimized linear layers. Notably, Our method demonstrates better performance under the same inference MACs.


\end{document}